%% file: main.tex
\documentclass{article}

\usepackage[nonatbib,final]{neurips_2024}

\usepackage[utf8]{inputenc} % allow utf-8 input
\usepackage[T1]{fontenc}    % use 8-bit T1 fonts
\usepackage[%
  unicode,
  linktocpage,
  backref=page,
]{hyperref}
\usepackage{url}
\usepackage{booktabs}       % professional-quality tables
\usepackage{amsfonts}       % blackboard math symbols
\usepackage{nicefrac}       % compact symbols for 1/2, etc.
\usepackage{microtype}      % microtypography
\usepackage{xcolor}         % colors

\usepackage{bbm}
\usepackage{apacite}
\usepackage[round]{natbib}

\usepackage{enumitem}
\usepackage{amsfonts}
\usepackage{setspace}
\usepackage{amsmath}
\usepackage{url}
\usepackage{graphicx}
\usepackage{booktabs}
\usepackage{multirow}
\usepackage{wrapfig}
\usepackage{caption}
\usepackage{subcaption}
\usepackage{soul}
\usepackage{dsfont}
\usepackage{todonotes}
\usepackage{rotating}

\usepackage{xspace} % for string command blanks handling
\usepackage{tikz}
\usetikzlibrary{bayesnet}
\usetikzlibrary{arrows}
\usepackage{color}
\usetikzlibrary{backgrounds}
\input{math_commands} % Need to copy this file
\usepackage{amsthm}

\usepackage{mathtools}

\AtBeginDocument{%
    \renewcommand*{\PrintBackRefs}[1]{}%
}
\let \backreforig \backref
\renewcommand*{\backref}[1]{[Referenced on page \backreforig{#1}]}

\definecolor{ForestGreen}{HTML}{009a55}
\definecolor{BrickRed}{HTML}{b6311e}
\definecolor{SunYellow}{HTML}{feec71}
\definecolor{DeepBlue}{HTML}{586e8b}
\definecolor{LegendaryBlue}{HTML}{3b7ab5}
\definecolor{LegendaryOrange}{HTML}{ff8225}
\definecolor{LegendaryGreen}{HTML}{3caf50}
\definecolor{LegendaryPink}{HTML}{fc7fbd}
\definecolor{VennOrange}{HTML}{ec7c30}
\definecolor{VennBlue}{HTML}{2e5496}
\definecolor{VennGreen}{HTML}{538235}
\usepackage[export]{adjustbox}

\makeatletter
\newcommand*{\defeq}{\mathrel{\rlap{%
                     \raisebox{0.3ex}{$\m@th\cdot$}}%
                     \raisebox{-0.3ex}{$\m@th\cdot$}}%
                     =}
\makeatother

\newcommand{\indep}{\perp \!\!\! \perp}

\newcommand{\method}{SCBM\xspace}
\title{Stochastic Concept Bottleneck Models} %: Modeling Concept Correlations in Concept Bottleneck Models

\author{% TODO: Maybe use official NeurIPS way of denoting authors, see commented-out below
  Moritz Vandenhirtz\thanks{Equal contribution. Correspondence to \texttt{\{moritz.vandenhirtz,slaguna\}@inf.ethz.ch}},  Sonia Laguna\footnotemark[1], Ričards Marcinkevičs, Julia E.~Vogt \\
  Department of Computer Science\\
  ETH Zurich\\
  Switzerland \\
}
\begin{document}

\maketitle

\begin{abstract}
Concept Bottleneck Models (CBMs) have emerged as a promising interpretable method whose final prediction is based on intermediate, human-understandable concepts rather than the raw input. 
Through time-consuming manual interventions, a user can correct wrongly predicted concept values to enhance the model's downstream performance.
We propose \emph{Stochastic Concept Bottleneck Models} (SCBMs), a novel approach that models concept dependencies. In SCBMs, a single-concept intervention affects all correlated concepts, thereby improving intervention effectiveness. Unlike previous approaches that model the concept relations via an autoregressive structure, we introduce an explicit, distributional parameterization that allows SCBMs to retain the CBMs' efficient training and inference procedure. 
Additionally, we leverage the parameterization to derive an effective intervention strategy based on the confidence region.
We show empirically on synthetic tabular and natural image datasets that our approach improves intervention effectiveness significantly. Notably, we showcase the versatility and usability of SCBMs by examining a setting with CLIP-inferred concepts, alleviating the need for manual concept annotations.
\end{abstract}
\setcounter{footnote}{0} 

\input{Sections/introduction}
\input{Sections/related_work}

\input{Sections/method}

\input{Sections/experiments}
\input{Sections/results}
\input{Sections/conclusion}

% \section{Preparing PDF files}

% Please prepare submission files with paper size ``US Letter,'' and not, for
% example, ``A4.''

% Fonts were the main cause of problems in the past years. Your PDF file must only
% contain Type 1 or Embedded TrueType fonts. Here are a few instructions to
% achieve this.

% \begin{itemize}

% \item You should directly generate PDF files using \verb+pdflatex+.

% \item You can check which fonts a PDF files uses.  In Acrobat Reader, select the
%   menu Files$>$Document Properties$>$Fonts and select Show All Fonts. You can
%   also use the program \verb+pdffonts+ which comes with \verb+xpdf+ and is
%   available out-of-the-box on most Linux machines.

% \item \verb+xfig+ "patterned" shapes are implemented with bitmap fonts.  Use
%   "solid" shapes instead.

% \item The \verb+\bbold+ package almost always uses bitmap fonts.  You should use
%   the equivalent AMS Fonts:
% \begin{verbatim}
%    \usepackage{amsfonts}
% \end{verbatim}
% followed by, e.g., \verb+\mathbb{R}+, \verb+\mathbb{N}+, or \verb+\mathbb{C}+
% for $\mathbb{R}$, $\mathbb{N}$ or $\mathbb{C}$.  You can also use the following
% workaround for reals, natural and complex:
% \begin{verbatim}
%    \newcommand{\RR}{I\!\!R} %real numbers
%    \newcommand{\Nat}{I\!\!N} %natural numbers
%    \newcommand{\CC}{I\!\!\!\!C} %complex numbers
% \end{verbatim}
% Note that \verb+amsfonts+ is automatically loaded by the \verb+amssymb+ package.

% \end{itemize}

% If your file contains type 3 fonts or non embedded TrueType fonts, we will ask
% you to fix it.

\begin{ack}
We thank Alexander Marx for the insightful discussions.
MV and SL are supported by the Swiss State Secretariat for Education, Research, and Innovation (SERI) under contract number MB22.00047. RM is supported by the SNSF grant \#320038189096.
\end{ack}

\newpage
\bibliographystyle{apacite}
\bibliography{bibliography}

%%%%%%%%%%%%%%%%%%%%%%%%%%%%%%%%%%%%%%%%%%%%%%%%%%%%%%%%%%%%

\newpage \appendix
\input{Sections/appendix}

\newpage
\section*{NeurIPS Paper Checklist}

\begin{enumerate}

\item {\bf Claims}
    \item[] Question: Do the main claims made in the abstract and introduction accurately reflect the paper's contributions and scope?
    \item[] Answer: \answerYes{} % Replace by \answerYes{}, \answerNo{}, or \answerNA{}.
    \item[] Justification: Claims are supported by evidence in the Results section and Appendix.

\item {\bf Limitations}
    \item[] Question: Does the paper discuss the limitations of the work performed by the authors?
    \item[] Answer: \answerYes{} % Replace by \answerYes{}, \answerNo{}, or \answerNA{}.
    \item[] Justification: Yes, we have a Limitations \& Future Work paragraph at the end of the conclusion. 

\item {\bf Theory Assumptions and Proofs}
    \item[] Question: For each theoretical result, does the paper provide the full set of assumptions and a complete (and correct) proof?
    \item[] Answer: \answerYes{} % Replace by \answerYes{}, \answerNo{}, or \answerNA{}.
    \item[] Justification: We provide derivations of the method's theoretical foundations (detailed up to an acceptable degree of expected math knowledge) in the Method section.

    \item {\bf Experimental Result Reproducibility}
    \item[] Question: Does the paper fully disclose all the information needed to reproduce the main experimental results of the paper to the extent that it affects the main claims and/or conclusions of the paper (regardless of whether the code and data are provided or not)?
    \item[] Answer: \answerYes{} % Replace by \answerYes{}, \answerNo{}, or \answerNA{}.
    \item[] Justification: We disclose hyperparameters in the main text and Appendix. We also offer the code for reproducibility in case any information is missing.

\item {\bf Open access to data and code}
    \item[] Question: Does the paper provide open access to the data and code, with sufficient instructions to faithfully reproduce the main experimental results, as described in supplemental material?
    \item[] Answer: \answerYes{} % Replace by \answerYes{}, \answerNo{}, or \answerNA{}.
    \item[] Justification: We have released an anonymized version of the repository.

\item {\bf Experimental Setting/Details}
    \item[] Question: Does the paper specify all the training and test details (e.g., data splits, hyperparameters, how they were chosen, type of optimizer, etc.) necessary to understand the results?
    \item[] Answer: \answerYes{} % Replace by \answerYes{}, \answerNo{}, or \answerNA{}.
    \item[] Justification: See Question 4.

\item {\bf Experiment Statistical Significance}
    \item[] Question: Does the paper report error bars suitably and correctly defined or other appropriate information about the statistical significance of the experiments?
    \item[] Answer: \answerYes{} % Replace by \answerYes{}, \answerNo{}, or \answerNA{}.
    \item[] Justification: We provide error bars in all experiments as we believe this to be of utmost importance to reproducible research. For the Appendix, we have reduced the number of seeds and/or experiment size to save computational resources for the environment's sake.

\item {\bf Experiments Compute Resources}
    \item[] Question: For each experiment, does the paper provide sufficient information on the computer resources (type of compute workers, memory, time of execution) needed to reproduce the experiments?
    \item[] Answer: \answerYes{} % Replace by \answerYes{}, \answerNo{}, or \answerNA{}.
    \item[] Justification: See Appendix
    
\item {\bf Code Of Ethics}
    \item[] Question: Does the research conducted in the paper conform, in every respect, with the NeurIPS Code of Ethics \url{https://neurips.cc/public/EthicsGuidelines}?
    \item[] Answer: \answerYes{} % Replace by \answerYes{}, \answerNo{}, or \answerNA{}.
    \item[] Justification: The code of ethics was followed.

\item {\bf Broader Impacts}
    \item[] Question: Does the paper discuss both potential positive societal impacts and negative societal impacts of the work performed?
    \item[] Answer: \answerNA{} % Replace by \answerYes{}, \answerNo{}, or \answerNA{}.
    \item[] Justification: Given the more foundational work of this paper, there is not a direct negative influence that the authors can think of that might arise from this work specifically.
    
\item {\bf Safeguards}
    \item[] Question: Does the paper describe safeguards that have been put in place for responsible release of data or models that have a high risk for misuse (e.g., pretrained language models, image generators, or scraped datasets)?
    \item[] Answer: \answerNA{} % Replace by \answerYes{}, \answerNo{}, or \answerNA{}.
    \item[] Justification: To the best of our knowledge, our work does not have high risk for misuse.

\item {\bf Licenses for existing assets}
    \item[] Question: Are the creators or original owners of assets (e.g., code, data, models), used in the paper, properly credited and are the license and terms of use explicitly mentioned and properly respected?
    \item[] Answer: \answerYes{} % Replace by \answerYes{}, \answerNo{}, or \answerNA{}.
    \item[] Justification: Licenses for all used datasets were clearly stated.

\item {\bf New Assets}
    \item[] Question: Are new assets introduced in the paper well documented and is the documentation provided alongside the assets?
    \item[] Answer: \answerYes{} % Replace by \answerYes{}, \answerNo{}, or \answerNA{}.
    \item[] Justification: In the Appendix, the data generating mechanism is clearly stated for the introduced synthetic dataset. Additionally, the new method is described in detail.

\item {\bf Crowdsourcing and Research with Human Subjects}
    \item[] Question: For crowdsourcing experiments and research with human subjects, does the paper include the full text of instructions given to participants and screenshots, if applicable, as well as details about compensation (if any)? 
    \item[] Answer: \answerNA{} % Replace by \answerYes{}, \answerNo{}, or \answerNA{}.
    \item[] Justification: The paper does not involve crowdsourcing nor research with human subjects.

\item {\bf Institutional Review Board (IRB) Approvals or Equivalent for Research with Human Subjects}
    \item[] Question: Does the paper describe potential risks incurred by study participants, whether such risks were disclosed to the subjects, and whether Institutional Review Board (IRB) approvals (or an equivalent approval/review based on the requirements of your country or institution) were obtained?
    \item[] Answer: \answerNA{} % Replace by \answerYes{}, \answerNo{}, or \answerNA{}.
    \item[] Justification: The paper does not involve crowdsourcing nor research with human subjects.

\end{enumerate}

\end{document}

%% file: math_commands.tex
\usepackage{amsmath,amsfonts,bm}

\def\eqref#1{equation~\ref{#1}}
\def\1{\bm{1}}

\def\vc{{\bm{c}}}

\def\vx{{\bm{x}}}

\def\mD{{\bm{D}}}

\def\mH{{\bm{H}}}
\def\mI{{\bm{I}}}

\def\mL{{\bm{L}}}

\def\mW{{\bm{W}}}

\def\mSigma{{\bm{\Sigma}}}

\DeclareMathAlphabet{\mathsfit}{\encodingdefault}{\sfdefault}{m}{sl}
\SetMathAlphabet{\mathsfit}{bold}{\encodingdefault}{\sfdefault}{bx}{n}

\DeclareMathOperator*{\argmax}{arg\,max}

\DeclareMathOperator*{\suchthat}{s.\!t.}

%% file: Sections/introduction.tex
\section{Introduction}

In today's world, machine learning plays a crucial role in making important decisions, from healthcare to finance and law. However, as these algorithms become more complex, understanding how they arrive at their decisions becomes increasingly challenging. This lack of interpretability is a significant concern, especially in situations where trustworthiness, transparency, and accountability are paramount~\citep{liptonMythosModelInterpretability2016, doshiRigorousScienceInterpretable2017}. Recent studies have focused on Concept Bottleneck Models (CBMs)~\citep{Koh2020, Havasi2022, Shin2023}, a class of models that predict human-understandable concepts upon which the final target prediction is based. CBMs offer interpretability since a user can inspect the predicted concept values to understand how the model arrives at its final target prediction. Moreover, if they disagree with a concept prediction, they can intervene by adjusting it to the right value, which in turn affects the target prediction.

\begin{figure}[t]
\centering
\includegraphics[width=\textwidth]{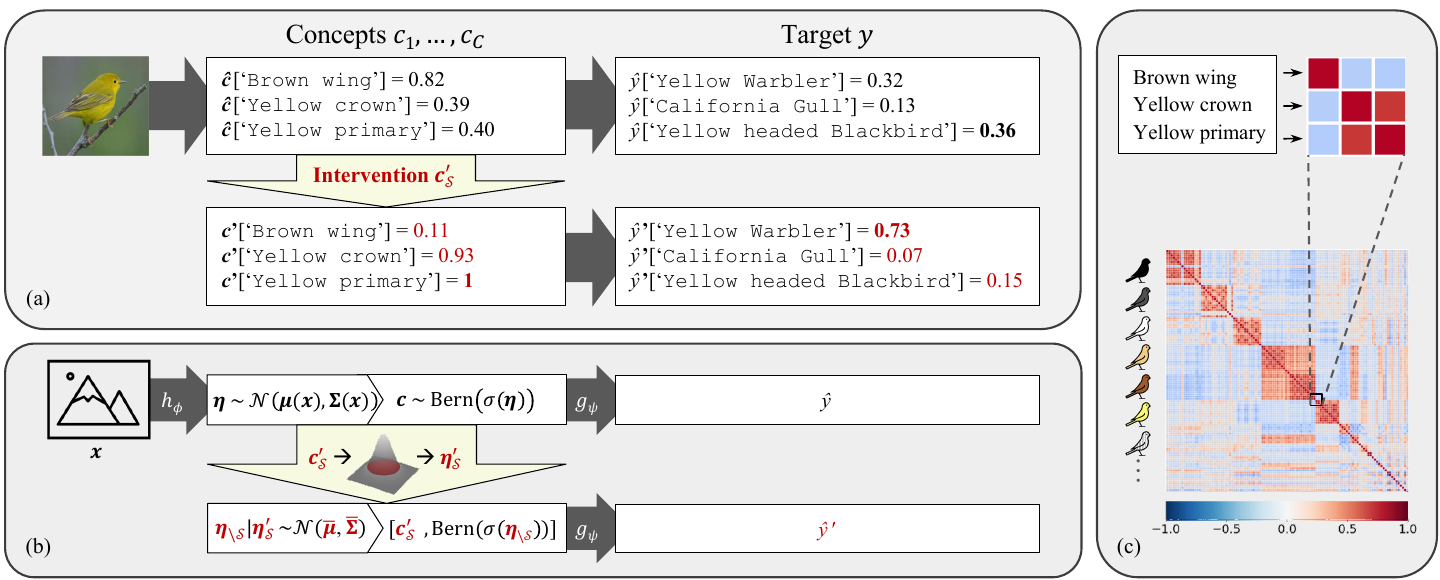}
\caption{Overview of the proposed method for the CUB dataset. (a) A user intervenes on the concept of `primary color: yellow'. Unlike CBMs, our method then uses this information to adjust the predicted probability of correlated concepts, thereby affecting the target prediction. (b) Schematic overview of the intervention procedure. A user's intervention $\vc'_{\mathcal{S}}$ is used to infer the logits $\boldsymbol{\eta}_{\setminus \mathcal{S}}$ of the remaining concepts. (c) Visualization of the learned global dependency structure as a correlation matrix for the 112 concepts of CUB~\citep{wah2011caltech}. Characterization of concepts on the left.}
\label{fig:SCBM}
\end{figure}

For example, consider the yellow warbler in Figure \ref{fig:SCBM} (a), where a user might notice that the binary concept `yellow primary color' is mispredicted. Upon this realization, they can intervene on the CBM by setting its value to $1$, which increases the probability of the class yellow warbler. This way of interacting allows any untrained user to engage with the model to increase its predictive performance.

However, if the user input is that the primary color is yellow, should not the likelihood of a yellow crown increase too? This adaptation would increase the predicted likelihood of the correct class even more, as yellow warblers are characterized by their fully yellow body. Currently, vanilla CBMs do not exhibit this behavior as they do not use the intervened-on concepts to update their remaining concept predictions. This indicates that they suboptimally adapt to the additional knowledge gained. To this end, we propose to extend the concept predictions with the modeling of their dependencies, as depicted in Figure~\ref{fig:SCBM}.

%Talk about how we can't intervene on 100 concepts and that there needs to be some scalable/efficient way of doing it. 
The proposed approach captures the concept dependencies by modeling the concept logits with a learnable non-diagonal normal distribution, which enables efficient, scalable computing of the effect of interventions on other concepts. By integrating concept correlations, we reduce the time and effort of having to laboriously intervene on many correlated variables and increase the efficacy of interventions on the downstream prediction. Thanks to the explicit distributional assumptions, the model is trained end-to-end, retaining the training and inference speed of classic CBMs as well as the benefits of training the concept and target predictor jointly.
Moreover, we show that our method excels when querying user interventions based on predicted concept uncertainty~\citep{Shin2023}, further highlighting the practical utility of our approach as such policies spare users from manually sifting through the concepts to identify necessary interventions. Lastly, based on the distributional concept parameterization, we propose a novel approach for computing dependency-aware interventions through the likelihood-based confidence region.

\paragraph{Contributions} 
This work contributes to the line of research on concept bottleneck models in several ways. (\emph{i})~We propose to capture and model concept dependencies with a multivariate normal distribution. (\emph{ii})~We derive a novel intervention strategy based on the confidence region of the normal distribution that incorporates concept correlations. 
%and empirically show its enhanced intervention effect. %
Using the learned concept dependencies during the intervention procedure allows for stronger interventional effectiveness. 
(\emph{iii})~We provide a thorough empirical assessment of the proposed method on synthetic tabular and natural image data. Additionally, we combine our method with concept discovery where we alleviate the need for annotations by using CLIP-inferred concepts.
In particular, we show the proposed method (a)~discovers meaningful, interpretable patterns in the form of concept dependencies, (b)~allows for fast, scalable inference, and (c)~outperforms related work with respect to intervention effectiveness thanks to the proposed concept modeling and intervention strategy.

%% file: Sections/related_work.tex
\section{Background \& Related Work}

Concept bottleneck models~\citep{Koh2020,Lampert2009,Kumar2009} are typically trained on data points $\left(\vx,\vc,y\right)$, comprising the covariates $\vx\in\mathcal{X}$, target $y\in\mathcal{Y}$, and $C$ annotated binary concepts $\vc\in\mathcal{C}$. Consider a neural network $f_{\boldsymbol{\theta}}$ parameterized by $\boldsymbol{\theta}$ and a slice $\left\langle g_{\boldsymbol{\psi}}, h_{\boldsymbol{\phi}}\right\rangle$ \citep{Leino2018} s.t.  $\hat{y}\defeq f_{\boldsymbol{\theta}}\left(\vx\right)=g_{\boldsymbol{\psi}}\left(h_{\boldsymbol{\phi}}\left(\vx\right)\right)$. CBMs enforce a concept bottleneck $\boldsymbol{\hat{c}}\defeq h_{\boldsymbol{\phi}}(\vx)$ such that the model's final output depends on the covariates $\vx$ solely through the predicted concepts $\boldsymbol{\hat{c}}$. 

While \citet{Koh2020} propose the \emph{soft} CBM, where the concept logits parameterize the bottleneck, \citet{Havasi2022} argue that such a representation leads to leakage, where additional unwanted information in the concept representation is used to predict the target~\citep{Margeloiu2021, Mahinpei2021}. Thus, they parameterize the bottleneck by binarized concept predictions and call it the \emph{hard} CBM. Then, \citet{Havasi2022} equip the hard CBM with an autoregressive structure of the form $c_i|\vx, \vc_{<i}$, which is supposed to learn the concept dependencies. As such, the implicit autoregressive modeling of concept dependencies by \citet{Havasi2022} is the most related to the current work. Complementary to our work, \citet{heidemann2023concept} analyze how a CBM's performance is affected by concept correlations. Unlike approaches that restrict the bottleneck to prevent leakage, Concept Embedding Models (CEM)~\citep{espinosa2022concept} represent each concept with an embedding vector from which the concept probabilities can be inferred. \citet{kim2023probabilistic} model the embedding with a normal distribution, assuming a diagonal covariance matrix, which prevents them from capturing concept dependencies. Therefore, their intervention performance is not expected to differ from that of CEMs. Recent works explored how a CBM-like structure can be enforced even without a concept-annotated training set. \citet{Yuksekgonul2023} transform a pre-trained model into a CBM via a concept bank from concept activation vectors and multimodal models~\citep{Kim2018}, while \citet{oikarinen2023label} query GPT-3~\citep{brown2020language} for the concept set $\mathcal{C}$ and assign the values of the concept activations to each datapoint $\vx$ with CLIP~\citep{radford2021learning} similarities. Similarly, \citet{panousis2023sparse} uses CLIP to probabilistically discover a sparse set of concepts for each input, which could be used in our model for a fully probabilistic pipeline. Lastly, \citet{marcinkevivcs2024beyond} instead relax the need for a concept labeled training set to a smaller validation set by fine-tuning a pre-trained model. 

Intervenability~\citep{marcinkevivcs2024beyond} is a crucial element of CBMs as it allows the user to correct wrongly predicted concepts $\boldsymbol{\hat{c}}$ to $\vc'$, which in turn affects the target prediction of the model $\hat{y}'$. If multiple concepts are intervened on sequentially, the order of interventions is important. To this end, \citet{Sheth2022} and \citet{Shin2023} explore multiple policies according to which the order of concepts is determined. \citet{chauhan2023interactive} propose to combine predefined policies with learnable weighting parameters, while \citet{espinosa2024learning} learn the policy itself. Concurrently, \citet{singhi2024improving} learn a realignment module to align concept predictions.
\citet{Steinmann2023} argue that instance-specific interventions are costly and store previous interventions in a memory to automatically reapply them for similar data points. Lastly, \citet{Collins2023} explore the advantages of including uncertainty rather than treating humans as oracles.

Our work models concept dependencies by parameterizing the bottleneck with a distribution. In a similar vein, Variational Autoencoders~\citep{kingma2013auto} parameterize the bottleneck with a normal distribution to model and generate new data. Stochastic Segmentation Networks~\citep{monteiro2020stochastic} parameterize the logits of a segmentation map with a non-diagonal normal distribution to capture the spatial correlations of pixels and model the aleatoric uncertainty. The modeling of uncertainty with a distribution is also explored by Bayesian Neural Networks~\citep{neal2012bayesian} that learn a probability distribution over the neurons of a neural network.

%% file: Sections/method.tex
\section{Methods}
We propose Stochastic Concept Bottleneck Models\footnote{The code is available here: \url{https://github.com/mvandenhi/SCBM}.} (\method), a novel concept-based method that relaxes the implicit CBM assumption of independent concepts. \method captures the concept dependencies by learning their multivariate distribution. As a result, interventions become more effective and scalable, as a single intervention can influence multiple correlated concepts. A schematic overview of the proposed method is depicted in Figure~\ref{fig:SCBM} (b).

\subsection{Model Formulation}

To capture the concept dependencies, we model the concept logits $\boldsymbol{\eta}$ with a learned multivariate normal distribution. Modeling logits with a normal distribution has proven to be effective in the context of segmentation~\citep{monteiro2020stochastic}. While \citet{monteiro2020stochastic} use it to capture the spatial dependencies of pixels, we, instead, model the relations between concepts, where the properties of the normal distribution will prove useful. A neural network is trained to predict the distribution's parameters $\boldsymbol{\eta}\mid\vx \sim \mathcal{N}\left(\boldsymbol{\mu}(\vx)),\boldsymbol{\Sigma}(\vx)\right)$, where $\boldsymbol{\mu}(\vx) \in \mathbb{R}^C$, and $\boldsymbol{\Sigma}(\vx) \in \mathbb{R}^{C\times C}$. Thus, the traditional assumption of independent concepts $c_i \indep c_j \mid \vx, \ \  \forall i \neq j$ is relaxed to $c_i \indep c_j \mid \boldsymbol{\eta}, \ \ \forall i \neq j$, where the assumed normal distribution induces linear concept dependencies. The inductive bias of linearity is useful in practice as it is more robust to overfitting and computationally more scalable with respect to $C$ compared to its nonlinear alternative~\citep{Havasi2022}, as we will show in Section~\ref{sec:results}.

To learn the distribution, we minimize the negative log-likelihood
\begin{equation}
    -\log p(\vc \mid \vx)=-\log \int p(\vc \mid \boldsymbol{\eta}) p_{\boldsymbol{\phi}} (\boldsymbol{\eta} \mid \boldsymbol{x}) d \boldsymbol{\eta},
\end{equation}
    
where $\boldsymbol{\phi}$ are the parameters of a neural network that predicts the distribution $\boldsymbol{\eta}\mid\vx \sim \mathcal{N}\left(\boldsymbol{\mu}(\vx)),\boldsymbol{\Sigma}(\vx)\right)$. This integral is intractable due to the softmax operation applied in $p(\vc \mid \boldsymbol{\eta})$. Thus, the integral is approximated by $M$ Monte Carlo samples
\begin{equation}
-\log \int p(\vc \mid \boldsymbol{\eta}) p_{\boldsymbol{\phi}} (\boldsymbol{\eta} \mid \boldsymbol{x}) d \boldsymbol{\eta} \approx -\log \frac{1}{M} \sum_{m=1}^M p(\vc \mid \boldsymbol{\eta}^{(m)}), \quad \boldsymbol{\eta}^{(m)}\mid\vx \sim \mathcal{N}\left(\boldsymbol{\mu}(\vx)),\boldsymbol{\Sigma}(\vx)\right).
\end{equation}
In order to learn $\boldsymbol{\phi}$, we make use of the parameterization as normal distribution and employ the reparameterization trick $\boldsymbol{\eta}^{(m)}\mid\vx = \boldsymbol{\mu}(\vx) + \mathbf{L}(\vx) \boldsymbol{\epsilon}^{(m)}, \quad \mathbf{L}(\vx)\mathbf{L}(\vx)^T = \boldsymbol{\Sigma}(\vx), \quad \boldsymbol{\epsilon}^{(m)}\sim \mathcal{N}\left(\boldsymbol{0},\boldsymbol{I}\right)$ such that gradients can be computed with respect to the parameters.
Lastly, we incorporate the new relaxed conditional independence assumption 
\begin{equation}
\log p(\vc \mid \boldsymbol{\eta}) = \log \prod_{i=1}^C p(c_i \mid \eta_i) = \sum_{i=1}^C \log p(c_i \mid \eta_i),
\end{equation}
where $p(c_i \mid \eta_i)$ describes a Bernoulli distribution parameterized by the sigmoid-transformed logits $\sigma(\eta_i)$.
Combining the above considerations results in the following reformulation of the negative log-likelihood:
\begin{align}
\begin{split}
-\log p(\vc \mid \vx) \approx& -\log \frac{1}{M} \sum_{m=1}^M p(\vc \mid \boldsymbol{\eta}^{(m)})\\
\propto& -\log \sum_{m=1}^M \exp \sum_{i=1}^C \log p(c_i \mid \eta_i^{(m)})\\
=& -\log \sum_{m=1}^M \exp \sum_{i=1}^C \left[-\mathrm{BCE}(c_i, \sigma(\eta_i^{(m)})) \right],
\end{split}
\end{align}
where BCE stands for Binary Cross Entropy, and the {\fontfamily{lmr}\selectfont
logsumexp} trick is used for numerical stability.

The distribution-based modeling procedure allows for efficient sampling, thus, enabling \method to train concept and target predictors jointly, sequentially, or independently. In contrast, the autoregressive alternative~\citep{Havasi2022} requires independent training due to the computational complexity. We adopt a joint training scheme to obtain the benefits of end-to-end learning where concept and target predictors can adjust to each other.
To prevent leakage, we follow \citet{Havasi2022} and train the model with the hard $\{0,1\}$ concept values as bottleneck rather than the logits used in the original CBM~\citep{Koh2020}. To this end, we employ the straight-through Gumbel-Softmax trick~\citep{Gumbel, concrete} that approximates Bernoulli samples while being differentiable. The target predictor $g_{\boldsymbol{\psi}}$ is then learned by minimizing the negative log-likelihood 
\begin{align}
\begin{split}
-\log p(y \mid \vx) =& - \log \sum_{\vc \in \mathcal{C}} p_{\boldsymbol{\psi}}(y \mid \vc)p(\vc \mid \vx)  \\
\approx& - \log \frac{1}{M} \sum_{m=1}^M p_{\boldsymbol{\psi}}(y \mid \vc^{(m)}), \qquad \vc^{(m)} \sim p(\vc \mid \vx).
\end{split}
\end{align}
Lastly, the learned dependencies are regularized by following Occam's razor and to prevent overfitting. We take inspiration from the Graphical Lasso~\citep{friedman2008sparse} and penalize the off-diagonal elements of the precision matrix $\boldsymbol{\Sigma}^{-1}$.

By combining concept, target, and precision loss with weighting factors $\lambda_1$ and $\lambda_2$, we arrive at the final loss function
\begin{equation}
\label{eq:loss}
    -\log \sum_{m=1}^M \exp \sum_{i=1}^C -\mathrm{BCE}\left(c_i, \sigma (\eta_i^{(m)})\right) + \lambda_1 \mathrm{CE}\left( y, 
    \frac{1}{M} \sum_{m=1}^M g_{\boldsymbol{\psi}}(\vc^{(m)})
\right) + \lambda_2 \sum_{i\neq j} \boldsymbol{\Sigma}(\vx)^{-1}_{i,j}.
\end{equation}

\subsection{Covariance Learning}
The introduced amortized covariance matrix $\boldsymbol{\Sigma}(\vx)$ provides the flexibility to tailor its predicted concept dependencies to each data point, making it adaptable to many data-generating mechanisms. For example, in the commonly used CUB~\citep{wah2011caltech, Koh2020}, it can learn the class-wise concept structure present in the dataset. 
The explicit dependency representation inferred by the learned covariance matrix is useful as it provides insights into the learned correlations among the concepts, which is important for understanding and interpreting the model behavior.

%However, using an amortized covariance matrix sacrifices a unified concept structure at the dataset level, limiting interpretability by precluding a single, comprehensive covariance matrix for the entire dataset.
However, an amortized covariance matrix comes at the price of not being able to visualize and interpret a unified concept structure on a dataset level. Depending on the need of the application, such a global structure might be preferable. Thus, we propose a variation of \method, where the covariance matrix is not \emph{amortized} ($\boldsymbol{\Sigma}(\vx)$), but learned \emph{globally} ($\boldsymbol{\Sigma}$). An example of the global concept structure learned on CUB is shown in Figure~\ref{fig:SCBM} (c). This variation has the inductive bias of assuming a constant covariance matrix, whose utility depends on the underlying data-generating mechanism. We recommend using the more flexible, amortized version by default and only utilizing a global covariance if the strong assumption of fixed dependencies is reasonable. We will explore this empirically in more detail in Section~\ref{sec:results}.

% Make sure it's all somewhere in the text: Normal distribution useful because easy conditioning, reparameterization trick for joint training (as opposed to havasi -> "which is not feasible if each concept is modelled sequentially), quick inference (as opposed to havasi), Explicit representation of learned covariance
% Normal distribution -> More interpretable /truthful probability estimates

\subsection{Interventions}
\label{subsec:int}

A distinguishing property of CBM-like methods is the user's capacity to correct wrongly predicted concepts, which in turn affects the target prediction~\citep{marcinkevivcs2024beyond}. For a big concept set, this intervention procedure can become quite laborious as a user has to inspect and manually intervene on each concept separately. SCBMs are designed to alleviate this need by utilizing the learned concept dependencies such that a single intervention affects all related concepts as modeled by the multivariate normal distribution.

The parameterization as a multivariate normal distribution allows for a quick, scalable intervention procedure. Given a set $\mathcal{S}\subset \{1,\ldots,C\}$ of concept interventions, the effect on the remaining concepts $\vc_{\setminus \mathcal{S}}$ is computed via their logits $\boldsymbol{\eta}_{\setminus \mathcal{S}}$ by conditioning on the intervention logits $\boldsymbol{\eta}_{\mathcal{S}}'$, utilizing the known properties of the normal distribution
\begin{align}
\label{eq:cond}
\begin{split}
\boldsymbol{\eta}_{\setminus \mathcal{S}} \mid \vx, \boldsymbol{\eta}'_{\mathcal{S}} &\sim \mathcal{N}\left(\boldsymbol{\bar\mu}(\vx), \boldsymbol{\overline{\Sigma}}(\vx)\right), \\
\boldsymbol{\bar\mu} &= \boldsymbol{\mu}_{\setminus \mathcal{S}} + \boldsymbol{\Sigma}_{\setminus \mathcal{S},\mathcal{S}} \boldsymbol{\Sigma}_{\mathcal{S},\mathcal{S}}^{-1}(\boldsymbol{\eta}'_{\mathcal{S}} - \boldsymbol{\mu}_{\mathcal{S}}), \\
\boldsymbol{\overline{\Sigma}} &= \boldsymbol{\Sigma}_{\setminus \mathcal{S},\setminus \mathcal{S}} - \boldsymbol{\Sigma}_{\setminus \mathcal{S},\mathcal{S}} \boldsymbol{\Sigma}_{\mathcal{S},\mathcal{S}}^{-1} \boldsymbol{\Sigma}_{\mathcal{S},\setminus \mathcal{S}}.
\end{split}
\end{align}
In standard CBMs, an intervention affects only the concepts on which the user intervenes. As such, \citet{Koh2020} set $\eta_i'$ to the 5th percentile of the training distribution if $c_i = 0$ and the 95th percentile if $c_i = 1$. While this strategy is effective for SCBMs too, see Appendix~\ref{app:interv_strat}, the modeling of the concept dependencies warrants a more thorough analysis of the \textit{intervention strategy}. We present two desiderata, which our intervention strategy should fulfill.
\vspace{-0.2cm}
\begin{enumerate}[label=\textit{\roman*)}]
    \item \textit{$p(c_i \mid \eta_i') \geq p(c_i \mid \mu_i)$}\vspace{0.1cm}\\
    The likelihood of the intervened-on concept $c_i$ should always increase after the intervention. If SCBMs used the same strategy as CBMs, it could happen that the initially predicted $\mu_i$ was more extreme than the selected training percentile. Then, the interventional shift $\eta'_i - \mu_i$ in Eq.~\ref{eq:cond} would point in the wrong direction. This would cause $\boldsymbol{\eta}_{\setminus \mathcal{S}}$ to shift incorrectly.
    \item \textit{$|\eta_i' - \mu_i|$ \text{should not be ``too large''}.}\vspace{0.1cm}\\
    We posit that the interventional shift should stay within a reasonable range of values. Otherwise, the effect on $\eta_{\setminus \mathcal{S}}$ would be unreasonably large such that the predicted $\boldsymbol{\mu}_{\setminus \mathcal{S}}$ would be completely disregarded.
\end{enumerate}
\vspace{-0.2cm}
To fulfill these desiderata, we take advantage of the explicit distributional representation: the likelihood-based confidence region of $\mu_i$ provides a natural way of specifying the region of possible $\boldsymbol{\eta}_{\mathcal{S}}'$ that fulfill our desiderata. Informally, a confidence region captures the region of plausible values for a parameter of a distribution.
Note that the confidence region takes concept dependencies into account when describing the area of possible $\boldsymbol{\eta}_{\mathcal{S}}'$. To determine the specific point within this region, we search for the values $\boldsymbol{\eta}_{\mathcal{S}}'$, which maximize the log-likelihood of the known, intervened-on concepts $\mathbf{c}_{\mathcal{S}}$, implicitly focusing on concepts that the model predicts poorly:

\begin{align}
\begin{split}
    \boldsymbol{\eta}_{\mathcal{S}}' = \argmax_{\boldsymbol{\eta}_{\mathcal{S}}} &\log p(\vc_{\mathcal{S}} \mid \boldsymbol{\eta}_{\mathcal{S}}) \\
    \suchthat &-2\left(\log p(\boldsymbol{\eta}_{\mathcal{S}} \mid \boldsymbol{\mu}_{\mathcal{S}}, \boldsymbol{\Sigma}_{\mathcal{S},\mathcal{S}}) - \log p(\boldsymbol{\mu}_{\mathcal{S}} \mid \boldsymbol{\mu}_{\mathcal{S}}, \boldsymbol{\Sigma}_{\mathcal{S},\mathcal{S}})\right) \leq \chi^2_{d, 1-\alpha}\\
    &\eta_i' - \mu_i \geq 0 \text{ if } c_i = 1, \quad \forall i \in \mathcal{S} \\
    &\eta_i' - \mu_i \leq 0 \text{ if } c_i = 0, \quad \forall i \in \mathcal{S},
\end{split}
\end{align}

where $d = |\mathcal{S}|$. The first inequality describes the confidence region. It is based on the logarithm of the likelihood ratio, which, after multiplying with $-2$, asymptotically follows a $\chi^2$ distribution~\citep{silvey1975statistical}. The last two inequalities restrict the region to the desired direction.
Note that $\boldsymbol{\eta}_{\mathcal{S}}'$ is computed to determine the conditional effect of the interventions on $\boldsymbol{\eta}_{\setminus \mathcal{S}}$ using Equation~\ref{eq:cond}. When predicting $\hat{y}'$ under interventions, the logits $\boldsymbol{\eta}_{\setminus \mathcal{S}}$ are then used for sampling the binary concept values $\vc_{\setminus \mathcal{S}}$ while the intervened-on concepts $\vc'_{\mathcal{S}}$ are directly set to their known, binary value.

%Note that after having computed the conditional effect of $\boldsymbol{\eta}_{\mathcal{S}}'$ on $\boldsymbol{\eta}_{\setminus \mathcal{S}}$, the intervened concepts $\vc'_{\mathcal{S}}$ are set to their known binary value for predicting $\hat{y}$.

%% file: Sections/experiments.tex
\section{Experimental Setup}
\label{sec:exp_setup}
\paragraph{Datasets and Evaluation}
% Synthetic explanation
We perform experiments on a variety of datasets to showcase the validity of our method. Inspired by \citet{Marcinkevics2023}, we introduce a synthetic tabular dataset with a data-generating mechanism that contains fixed concept dependencies we can regulate. In particular, the concept logits $\boldsymbol{\eta}$ are sampled from a randomly initialized positive definite covariance matrix and generate $\vx$. Binary concept values $\vc$ are inferred from $\boldsymbol{\eta}$ and generate the target $y$. We refer to Appendix~\ref{sec:app:syn} for a more detailed description.

%CUB
As a natural image classification benchmark, we evaluate on the Caltech-UCSD Birds-200-2011 dataset \citep{wah2011caltech}, comprised of bird photographs from 200 distinct classes. It includes 112 concepts, such as wing color and beak shape, shared across the same class instances as revised in the original CBM work \citep{Koh2020}. Additionally, we explore another natural image classification task on CIFAR-10 \citep{krizhevsky2009learning} with 10 classes. To mitigate the concept annotations requirement,  the concepts are synthetically acquired in a similar fashion to the concept discovery literature. We adopt the 143 concept classes generated via GPT-3~\citep{brown2020language} in prior work \citep{oikarinen2023label}. To obtain the binary concept values, we use the CLIP model~\citep{radford2021learning} to compute the similarity between each instance of an image with the text embedding of a specific concept and compare it to the similarity of its negative counterpart, i.e. \emph{not} the concept. Appendix~\ref{app:nat_ima} contains further details about the natural image datasets.

To compare methods, we evaluate the model performance based on the concept and target accuracy. We compute test performance before and after intervening on an increasing number of concepts.
%We compute test performance prior to interventions, as well as the performance after intervening on an increasing number of ground truth concepts. 
The order of concepts in the intervention is determined by an uncertainty-based policy~\citep{Shin2023} that selects the concept whose predicted probability is closest to $0.5$. We also show results for a random policy in Appendix~\ref{sec:app:rand}. Additionally, we evaluate the calibration of the predicted concept uncertainties that are being used for the uncertainty-based policy, with the Brier score~\citep{brier1950verification} and the Expected Calibration Error~\citep{naeini2015obtaining,kumar2019verified}.

\paragraph{Baselines}
% Hard CBM, AR, CEM
We evaluate the performance of our method in comparison with state-of-the-art models. Namely, we focus on the vanilla concept bottleneck model (CBM) by \citet{Koh2020} in its \emph{hard} version~\citep{Havasi2022}, trained jointly using the straight-through Gumbel-Softmax trick~\citep{Gumbel, concrete}, as a sensical baseline to our binary modeling of concepts. 
Additionally, we explore the concept embedding model (CEM) by \citet{espinosa2022concept} that learns two concept embeddings, $\boldsymbol{\hat{c}}_i^+$ and $\boldsymbol{\hat{c}}_i^-$. These representations are used to predict the final concept probability with a learnable scoring function $\hat{p}_i = s(\boldsymbol{\hat{c}}_i^+, \boldsymbol{\hat{c}}_i^-) = \sigma(\mathbf{W}_s [\boldsymbol{\hat{c}}_i^+, \boldsymbol{\hat{c}}_i^-]^T + \mathbf{b}_s)$ and are then combined into a final concept embedding $\boldsymbol{\hat{c}}_i = (\hat{p}_i \boldsymbol{\hat{c}}_i^+ + (1 - \hat{p}_i) \boldsymbol{\hat{c}}_i^-)$ that is passed to the target predictor. Interventions are modeled by altering the concept probabilities $\hat{p}_i$. Note that \citet{espinosa2022concept} optimize for intervention performance during training, which we omit, to ensure a fair comparison where no method was explicitly trained for intervention performance.
Finally, we evaluate the autoregressive CBM structure proposed by \citet{Havasi2022}, where concept dependencies are learned with an autoregressive structure. Here, each concept $c_i$ is predicted with a separate MLP that takes as input a latent representation of the input $f_{\boldsymbol{\theta}}(\vx)$ and all previous concepts $c_1,...,c_{i-1}$. To obtain a good initialization of the autoregressive structure, it is pretrained for $50$ epochs.
As the Monte Carlo sampling from the autoregressive structure is time-consuming, the target predictor $g_{\boldsymbol{\psi}}$ is trained independently using the ground-truth concepts as input.
At intervention time, a normalized importance sampling algorithm is used to estimate the concept distribution.

\paragraph{Implementation Details}
The model architectures comprise a backbone for concept prediction followed by a linear layer as head for an interpretable target prediction. More details can be found in Appendix~\ref{app:imp_det}. To ensure the positive definiteness of the concept covariance matrix $\boldsymbol{\Sigma}$, we parameterize it via its Cholesky decomposition $\boldsymbol{\Sigma}=\mL \mL^{\top}$. Thus, we directly predict the lower triangular Cholesky matrix $\mL$. We will evaluate two options for SCBMs: using a \emph{global} ($\boldsymbol{\Sigma}$) or an \emph{amortized} covariance matrix $(\boldsymbol{\Sigma}(\vx))$. For the amortized version, we set the weighting terms $\lambda_1$ and $\lambda_2$ of Equation~\ref{eq:loss} to 1. For the global version, we initialize it with the estimated empirical covariance matrix and set $\lambda_2=0$, as we did not observe big differences when varying $\lambda_2$. In Appendix~\ref{app:reg_strength}, we provide an ablation study, demonstrating that SCBMs are not very sensitive to the choice of $\lambda_2$. At intervention time, we solve the optimization problem based on the $99\%$-confidence region with the SLSQP algorithm~\citep{kraft1988software}. In Appendix~\ref{app:conf_region_level}, we provide an ablation with different confidence levels. 

%% file: Sections/results.tex
\begin{table}
\caption{Test-set concept and target accuracy (\%) prior to interventions. Results are reported as averages and standard deviations of model performance across ten seeds. For each dataset and metric, the best-performing method is \textbf{bolded} and the runner-up is \underline{underlined}.}
\label{tab:test_performance}
% Made small by Ctrl+F -> (\$\\pm\$ \d+\.\d+), replace: \\small{$1}
\centering
\begin{tabular}{llcc}
\toprule
Dataset & Method & Concept Accuracy & Target Accuracy  \\
\midrule
 & Hard CBM & 61.42 \small{$\pm$ 0.07} & 58.38 \small{$\pm$ 0.39} \\
 & CEM & 61.42 \small{$\pm$ 0.12} & 58.01 \small{$\pm$ 0.49} \\
Synthetic & Autoregressive CBM & \underline{62.17} \small{$\pm$ 0.11} & \textbf{59.60} \small{$\pm$ 0.62} \\
 & Global SCBM & 61.57 \small{$\pm$ 0.05} & 58.39 \small{$\pm$ 0.53} \\
 & Amortized SCBM & \textbf{62.41} \small{$\pm$ 0.20} & \underline{58.96} \small{$\pm$ 0.38} \\
\midrule
 & Hard CBM & 94.97 \small{$\pm$ 0.07} & 67.72 \small{$\pm$ 0.57} \\
 & CEM & 95.12 \small{$\pm$ 0.07} & \underline{69.60} \small{$\pm$ 0.30} \\
CUB & Autoregressive CBM & \textbf{95.33} \small{$\pm$ 0.07} & 69.24 \small{$\pm$ 0.44} \\
 & Global SCBM & 94.99 \small{$\pm$ 0.09} & 68.19 \small{$\pm$ 0.63} \\
 & Amortized SCBM & \underline{95.22} \small{$\pm$ 0.09} & \textbf{69.87} \small{$\pm$ 0.56} \\
\midrule
 & Hard CBM & 85.51 \small{$\pm$ 0.04} & 69.73 \small{$\pm$ 0.29} \\
 & CEM & 85.12 \small{$\pm$ 0.14} & \textbf{72.24} \small{$\pm$ 0.33} \\
CIFAR-10 & Autoregressive CBM & 85.31 \small{$\pm$ 0.06} & 68.88 \small{$\pm$ 0.47} \\
 & Global SCBM & \underline{85.86} \small{$\pm$ 0.04} & 70.74 \small{$\pm$ 0.29} \\
 & Amortized SCBM & \textbf{86.00} \small{$\pm$ 0.03} & \underline{71.66} \small{$\pm$ 0.25} \\
\bottomrule
\end{tabular}
\end{table}

\section{Results\label{sec:results}}
\paragraph{Test performance}
\begin{wraptable}{r}{0.5\textwidth} % 'r' for right, and 0.6\textwidth for table width
\vspace{-0.4cm}
\caption{Relative time it takes for one epoch in the CUB dataset when training on the training set, or evaluating on the test set, respectively.}
\label{tab:time}
\centering
\begin{tabular}{lcc}
\toprule
Method & Training & Inference \\
\midrule
Hard CBM           & 5x               & 1x            \\
CEM      & 5x              & 1x              \\
Autoregressive CBM & 5x            & 15x            \\
Global SCBM         & 5x               & 1x             \\
Amortized SCBM      & 5x              & 1x            \\
\bottomrule
% Hard CBM           & 5.30x               & 1.00x            \\
% CEM      & 5.38x              & 1.00x              \\
% Autoregressive CBM & 5.29x            & 14.78x            \\
% Global SCBM         & 5.35x               & 1.01x             \\
% Amortized SCBM      & 5.39x              & 1.03x            \\
% autoregressive  & 11.66s & 32.57s  \\
% embedding & 11.85s & 2.21s \\
% hard & 11.68s & 2.20s  \\
% global & 11.78s & 2.23s \\
% amortized & 11.87s & 2.26s
% autoregressive  & 11.66283984184265s & 32.56752460002899s  \\
% embedding & 11.853167724609374s & 2.2121152877807617s \\
% hard & 11.67969057559967s & 2.202788805961609s  \\
% global & 11.782135272026062s & 2.231656813621521s \\
% amortized & 11.872103214263916s & 2.2602181673049926s
\vspace{-0.8cm}
\end{tabular}
\end{wraptable}
In Table \ref{tab:test_performance}, we report the results of the concept and target accuracy prior to interventions. Overall, SCBM performs on par with the baseline methods, with no clear outperforming or underperforming technique throughout the datasets. In Appendix~\ref{app:jaccard}, we show that other metrics lead to the same interpretation. This shows that the additional overhead of learning the concept dependencies does not negatively affect the predictive performance. We note that the amortized covariance variant consistently surpasses the globally learned matrix due to its ability to adjust the predicted concept dependency structure and uncertainty on an instance level.
%We note that the amortized covariance version consistently surpasses the globally learned matrix, which is due to its flexibility that helps in capturing the complexity of the per-instance data. 
On the other hand, the global variant offers a unified understanding of the concept correlations, an example of which is presented in Figure~\ref{fig:SCBM} (c). 
Notably, in CIFAR-10, even though the concept performance of CEM is the worst of all methods, it has the best target performance. This might suggest the presence of leakage in CEM's embeddings, as in CIFAR-10, the concept set alone is not sufficient to predict the target, and learning additional information might be useful. In Table~\ref{tab:time}, we show the time it takes for training and testing of the methods. It is evident that the autoregressive CBM of \citet{Havasi2022} suffers from a slow sampling process due to its autoregressive structure, while SCBMs retain the efficiency of CBMs.

\begin{figure*}[htbp]
    \centering
    %\textbf{Uncertainty-based Interventions}\par\medskip
    \begin{subfigure}[t]{0.32\linewidth} % Changed alignment to top
        \centering % Changed to centering instead of flushright for symmetry
        \includegraphics[height=3.3cm,keepaspectratio,right]{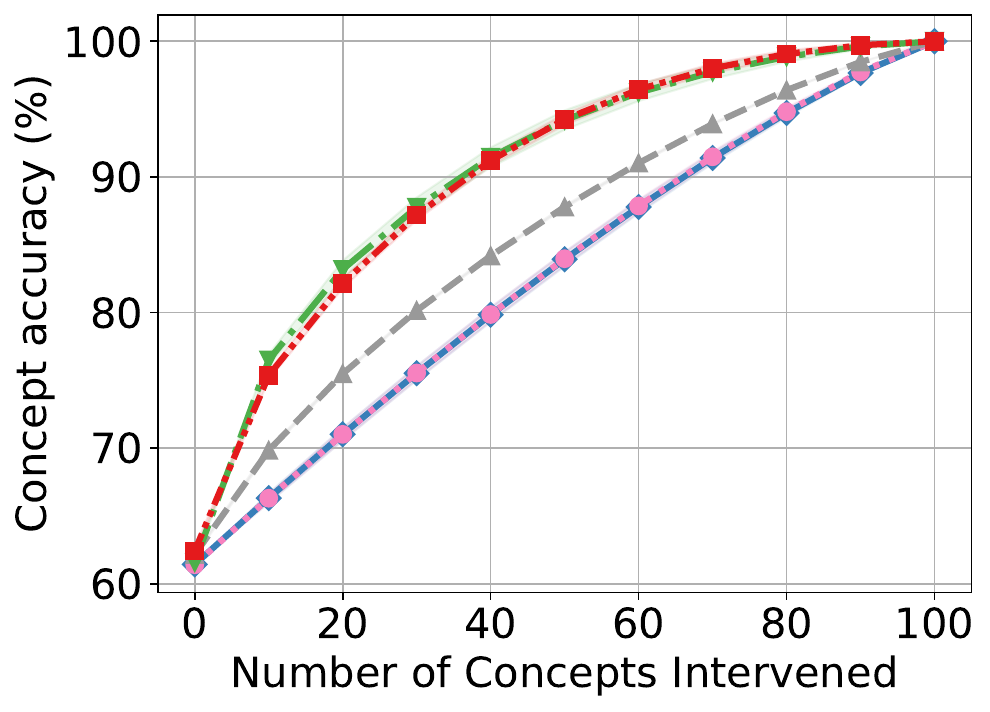}
        \includegraphics[height=3.3cm,keepaspectratio,right]{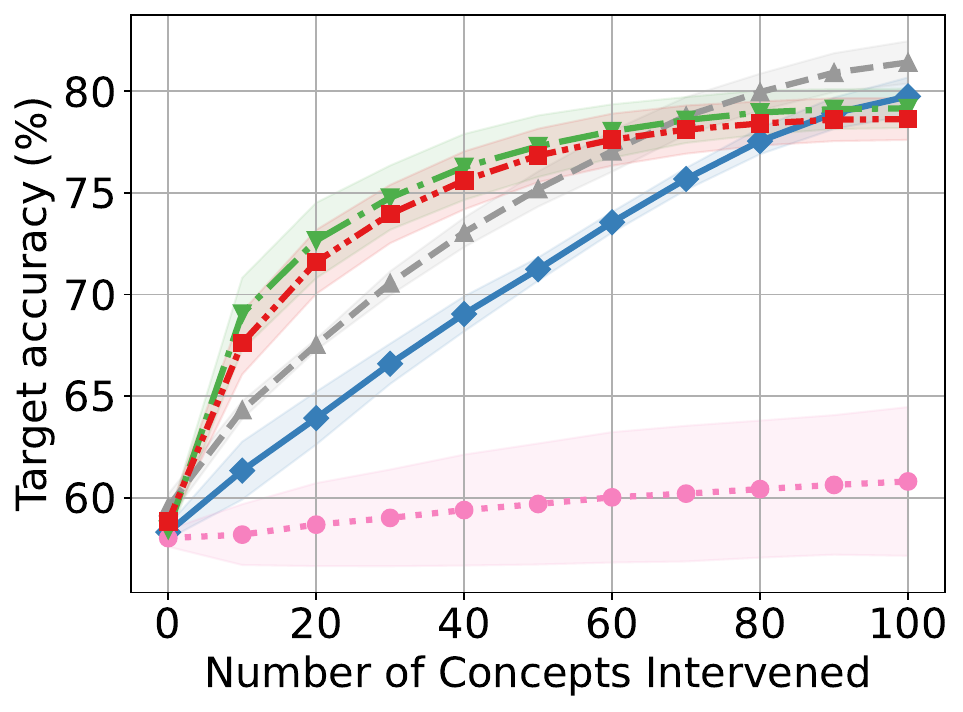}
        \caption{\footnotesize {Synthetic}}
    \end{subfigure}
    \begin{subfigure}[t]{0.32\linewidth} % Changed alignment to top
        \centering % Changed to centering instead of flushright for symmetry
        \includegraphics[height=3.3cm,keepaspectratio,right]{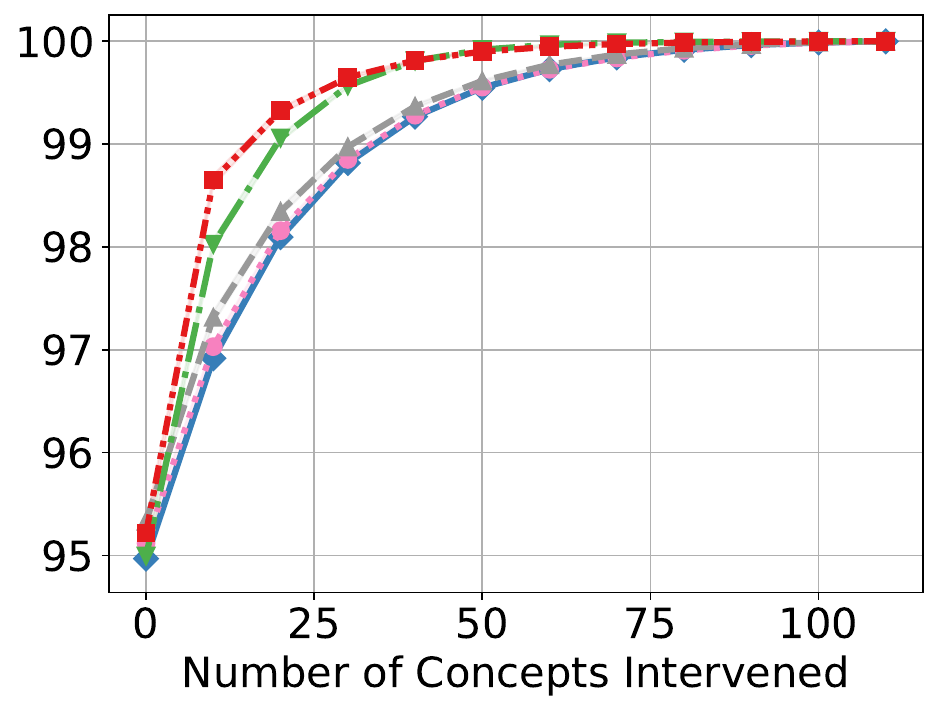}
        \includegraphics[height=3.3cm,keepaspectratio,right]{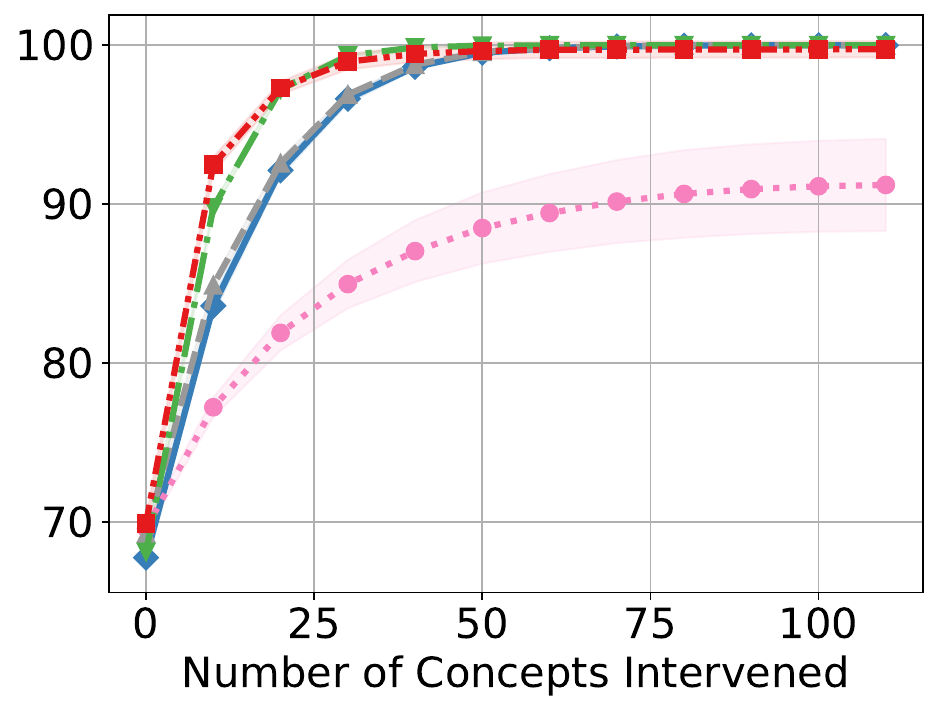}
        \caption{\footnotesize {CUB}}
    \end{subfigure}
    \begin{subfigure}[t]{0.32\linewidth} % Changed alignment to top
        \centering % Changed to centering instead of flushright for symmetry
        \includegraphics[height=3.3cm,keepaspectratio,right]{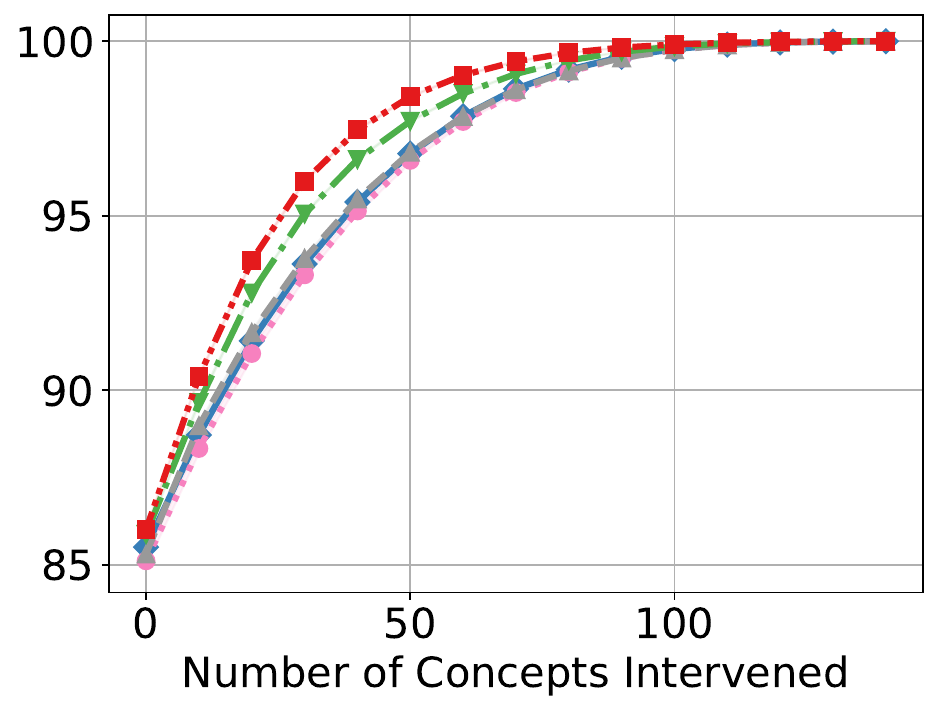}
        \includegraphics[height=3.3cm,keepaspectratio,right]{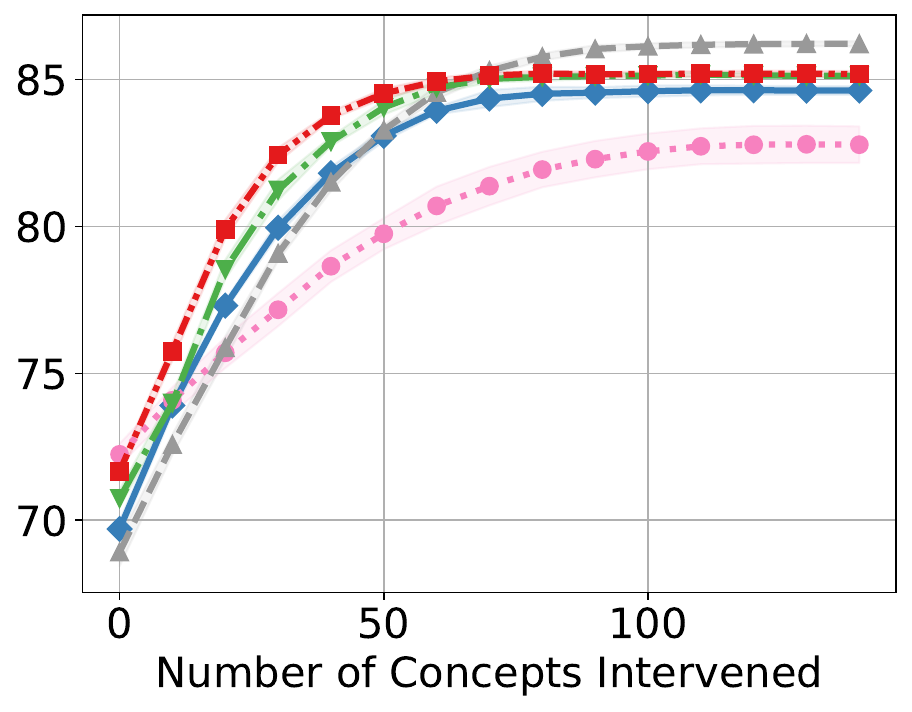}
        \caption{\footnotesize {CIFAR-10}}
    \end{subfigure}
    \includegraphics[width=0.9\linewidth]{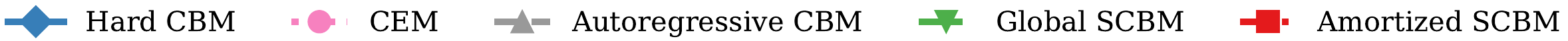}
    \caption{Performance after intervening on concepts in the order of highest predicted uncertainty. Concept and target accuracy (\%) are shown in the first and second rows, respectively. Results are reported as averages and standard deviations of model performance across ten seeds.}
    \label{inter_plots_uncert}
\end{figure*}
\vspace{-0.3cm}
\paragraph{Interventions} In this paragraph, we analyze the intervention performance of SCBMs and their baseline models, focusing on their effectiveness in modeling concept dependencies and improving target accuracy. Figure~\ref{inter_plots_uncert} shows the intervention curves across ten seeds, where the performance is measured based on the concept and target accuracy. The order of concepts to intervene on is determined by an uncertainty-based policy that makes use of the predicted probabilities. In Appendix~\ref{sec:app:rand}, we present the intervention performance if concepts were selected randomly. The intervention curves in the first row show that SCBMs are superior in modeling the concept dependencies, as evidenced by their significantly steeper intervention curves compared to the baseline methods. Furthermore, the second row of Figure~\ref{inter_plots_uncert} indicates that the strong concept modeling translates to a significant improvement in downstream performance, partly thanks to the intervention strategy introduced in Section \ref{subsec:int}. We note that especially for the most practical scenario of only a small number of interventions, SCBMs outperform their counterparts. Comparing the SCBM variants, the natural image datasets show an overall better intervention performance with the amortized covariance matrix, following the trend of Table \ref{tab:test_performance}, as it can capture the instance-wise correlation structure of the data. Only in the synthetic dataset, where the data-generating covariance matrix is fixed, does the global SCBM slightly outperform the amortized one. Thus, we advocate for the usage of the global variant only if the underlying assumption of a fixed covariance is reasonable. Lastly, the success of SCBMs on CIFAR-10, with CLIP-based concepts, shows our proposed method can work without human-annotated concepts. To strengthen this point and also showcase the scalability of our method, in Appendix~\ref{app:cifar100}, we provide results on CIFAR-100 with 892 concepts, where our SCBMs also strongly outperform baselines.

Analyzing the performance of the autoregressive CBM, which also captures concept dependencies, we observe that they expectedly have a better intervention performance than the hard vanilla CBM, which does not take correlations into account. However, it becomes evident that, compared to the concept performance of SCBMs, their autoregressive structure does not capture the dependencies to the full extent. This shows in the target accuracy, where they only match or outperform SCBMs towards the full set of intervened concepts. 
We attribute the better performance on the full intervention set to the independent training procedure utilized by autoregressive CBMs, which comes at the cost of lower test performance in CIFAR-10.
%The improvements towards the full intervention set occur because the autoregressive structure forces them to train the target predictor on the ground truth concepts rather than the predictions.
%Thus, when the concept interventions approach the full set, then an independently trained target predictor becomes optimal, at the cost of lower test performance in CIFAR-10. 
Arguably, in a realistic use-case, such a high number of instance-level interventions is not sensible, and if it were, SCBMs could also be trained independently.
Finally, the CEM shows reduced intervention performance as the expressive concept embeddings, which are prone to information leakage, seem to suboptimally adapt to the injected concept information.
%makes it harder to intervene on isolated concept information.
% Finally, it is relevant to note that unlike our SCBM, the autoregressive CBM is trained independently due to the computational complexity of joint training for their model. Hereafter, the target prediction has been trained with the ground truth concepts as input, which can lead to improved performance when a large number of concepts are intervened on, as seen towards the last interventions on synthetic data or CIFAR-10. However, in a realistic use-case scenario, such a high number of instance-level interventions is not sensically required. On the contrary, as our modeling allows for joint training, we learn a better representation of the target variable and, hence, result in better performance without interventions and a steeper increase in the practical range of number of intervened on concepts.  
%Across datasets, the autoregressive CBM that is enabled with connections to capture concept correlations and a weighted intervention sampling process performs better than the vanilla hard CBM that does not take correlations into account. 

\begin{figure}[htbp]
    \centering
    \begin{minipage}{0.65\textwidth}
        \captionof{table}{Test-set calibration (\%) of concept predictions. Results are reported as averages and standard deviations of model performance across ten seeds. For each dataset and metric, the best-performing method is \textbf{bolded} and the runner-up is \underline{underlined}. Lower is better.}
        \centering
        \begin{tabular}{llcc}
        \toprule
        Dataset & Method & Brier  & ECE  \\
        \midrule
         &  Hard CBM & 28.79  \small{$\pm$ 0.09}  & 22.38  \small{$\pm$ 0.15}  \\
         & CEM & 29.32  \small{$\pm$ 0.08}  & 23.55  \small{$\pm$ 0.09}  \\
        Synthetic &  Autoregressive CBM & \textbf{24.84}  \small{$\pm$ 0.32}  & \textbf{13.54}  \small{$\pm$ 0.49}  \\
         & Global SCBM & 27.73  \small{$\pm$ 0.09}  & 20.10 \small{$\pm$ 0.14}  \\
         & Amortized SCBM & \underline{25.58}  \small{$\pm$ 0.20} & \underline{15.57}  \small{$\pm$ 0.55}  \\
        \midrule
         &  Hard CBM & 3.93  \small{$\pm$ 0.05}  & 2.44  \small{$\pm$ 0.06}  \\
         & CEM & 4.04  \small{$\pm$ 0.05}  & 3.25  \small{$\pm$ 0.07}  \\
        CUB &  Autoregressive CBM & \underline{3.75}  \small{$\pm$ 0.05}  & 2.73  \small{$\pm$ 0.05}  \\
         & Global SCBM & 3.87  \small{$\pm$ 0.06}  & \underline{2.33}  \small{$\pm$ 0.09}  \\
         & Amortized SCBM & \textbf{3.64}  \small{$\pm$ 0.07}  & \textbf{1.85}  \small{$\pm$ 0.08}  \\
        \midrule
         &  Hard CBM & 10.42  \small{$\pm$ 0.05}  & 4.93  \small{$\pm$ 0.17}  \\
         & CEM & 11.06  \small{$\pm$ 0.16}  & 7.11  \small{$\pm$ 0.39}  \\
        CIFAR-10 &  Autoregressive CBM & 10.70 \small{$\pm$ 0.05}  & 6.07  \small{$\pm$ 0.10} \\
         & Global SCBM & \underline{9.95}  \small{$\pm$ 0.02}  & \underline{2.88}  \small{$\pm$ 0.11}  \\
         & Amortized SCBM & \textbf{9.84}  \small{$\pm$ 0.02}  & \textbf{2.22}  \small{$\pm$ 0.12}  \\
        \bottomrule
        \end{tabular}
        \label{tab:calibration_exp}
    \end{minipage}
    \hfill
    \begin{minipage}{0.3\textwidth}
    \vspace{-0.3cm}
        \caption{Intervention performance of SCBMs measured in concept and target accuracy (\%) on CUB for random and uncertainty-based policy.}
        \centering
        \includegraphics[width=\linewidth]{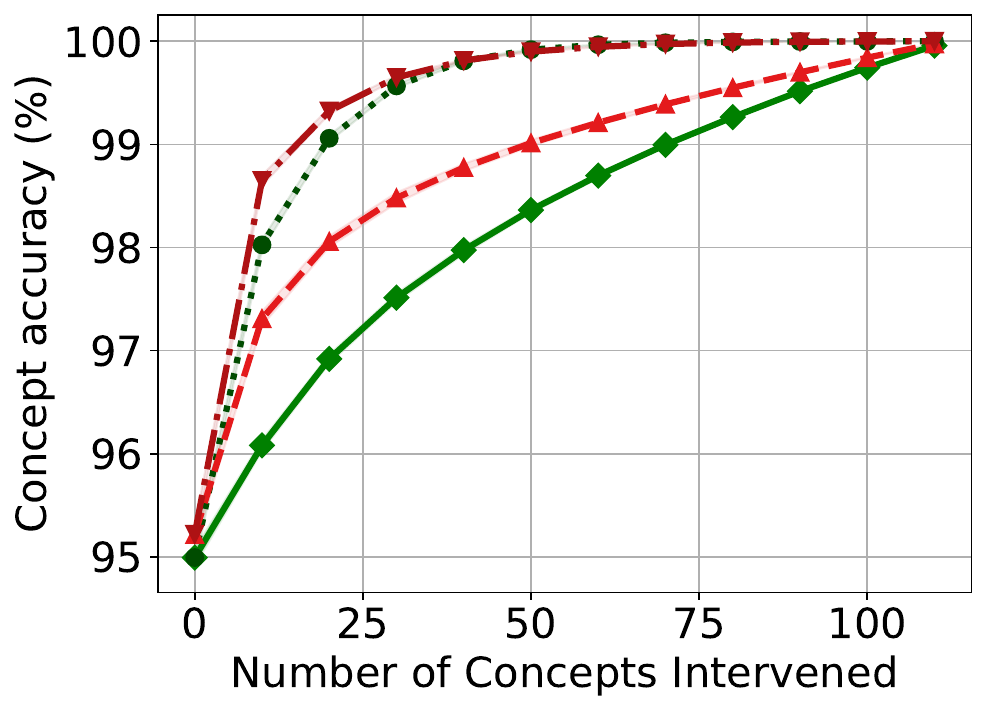}
    \includegraphics[width=\linewidth]{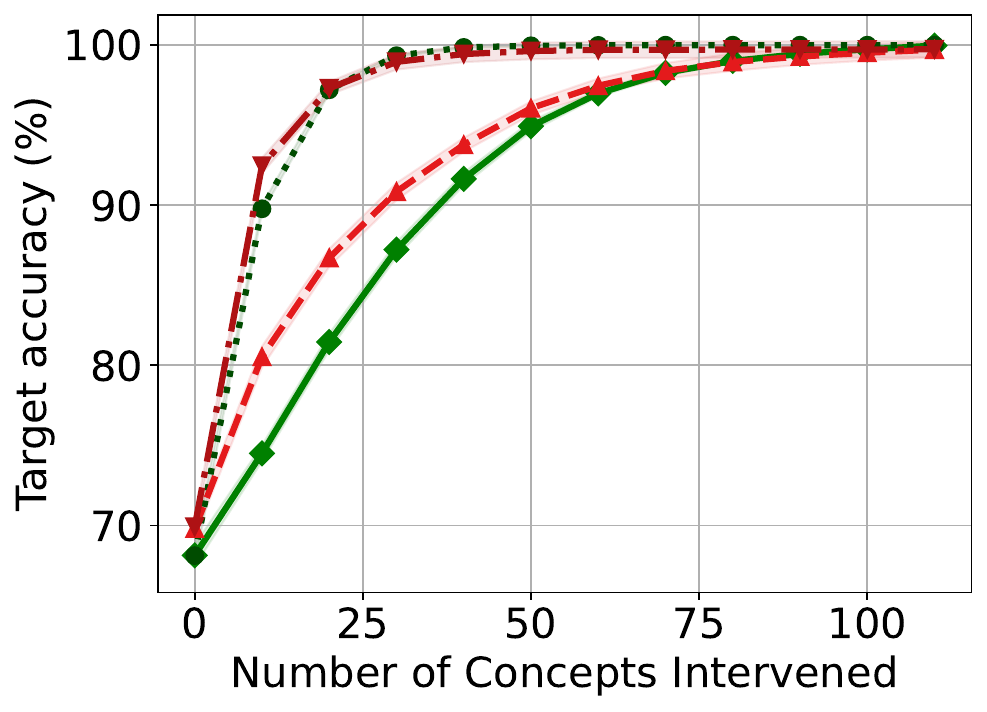}
    \includegraphics[width=1\linewidth]{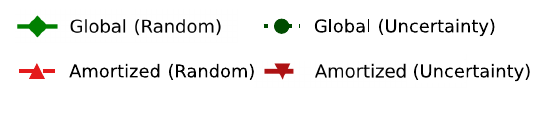}
    \vspace{-0.7cm}
        \label{fig:random_vs_unc}
    \end{minipage}
\end{figure}
\paragraph{Modeling the concept distribution}
A cornerstone of SCBMs is the explicit, distributional parameterization of concepts. This helps in understanding the data correlations and allows for visualization, as the example seen in Figure \ref{fig:SCBM} (c). The explicit probabilistic modeling results in improved concept uncertainty estimates compared to the baseline CBM counterparts, as shown in Table \ref{tab:calibration_exp}, where lower metrics imply better estimates. This proves useful for interventions, where the uncertainty estimates can be leveraged for the choice of concepts to intervene on, improving the target prediction more effectively and reducing the need for manual user inspection. 
In Figure \ref{fig:random_vs_unc}, we compare the performance of randomly intervening versus intervening based on the predicted uncertainty. We observe that there is a big gap between the two policies, indicating the usefulness of the estimated probabilities.
%Figure \ref{fig:random_vs_unc} shows on the CUB dataset how intervening based on the uncertainty estimates is more effective than selecting the concepts in a random fashion, hence, we focus on the former strategy through this work. The more meaningful uncertainty estimates enable our method to show a bigger gap between these two strategies. 
Nevertheless, note that intervening at random remains successful and supports the observations made in the previous paragraph, as shown in Appendix~\ref{sec:app:rand}.
%and compares similarly between baselines that the uncertainty-based policy. 

%% file: Sections/conclusion.tex
\section{Conclusion}

In this paper, we introduced SCBMs, a new concept-based method that models concept dependencies with a multivariate normal distribution. We proposed a novel, effective intervention strategy that takes concept correlations into account and is based on the confidence region inferred from the distributional parameterization. We showed that our modeling approach retains CBMs' training and inference speed, thus, being able to harness the benefits of end-to-end concept and target training. Additionally, the explicit parameterization offers the user a clearer understanding of the learned concept dependencies, providing deeper insights into how predictions and interventions are made. 
%With the proposed method, it is computationally affordable to train the concepts and target jointly, providing benefits in the target prediction with respect to the methods that model concept correlations in the state of the art, such as \citet{Havasi2022}, where only independent training is explored due to the elevated computational cost of joint training from sampling.
Empirically, we demonstrated that by modeling the concept dependencies, SCBMs offer a substantial improvement in intervention effectiveness, in concept as well as target accuracy, compared to related work. We showed that our method excels when iteratively intervening on the most uncertain concept predictions, sparing users from having to manually search through the concept set to identify necessary interventions. Additionally, our results indicate that learning the concept correlations does not decrease performance prior to interventions, in many cases even improving the performance over the baselines. Finally, the versatility of SCBMs is highlighted through their superior performance on CIFAR-10 and CIFAR-100, where concept values are CLIP-based rather than human-annotated.

\newpage
\paragraph{Limitations \& Future Work}
This work opens multiple new research avenues. A natural extension is to go beyond binary concepts, such as continuous domains with their corresponding adaptations of modeling the concept distribution. Additionally, addressing the quadratic memory complexity of the covariance matrix is essential for scaling to larger concept sets. Our proposed intervention strategy accounts for model uncertainty, but further research is needed to accommodate user uncertainty, as human interventions are not always the ground truth. This work allows the editing of the learned dependency structure by adjusting the entries of the predicted covariance matrix, which could be explored. Lastly, to model additional information and reduce leakage, \citet{Koh2020,Havasi2022} propose the adoption of a side channel. The complementary effectiveness of incorporating the side channel in the covariance structure could be explored in the context of SCBMs.

%% file: Sections/appendix.tex
\section{Dataset Details}
In this section, we provide additional details on the datasets that are being used in the experiments.
\subsection{Synthetic Data-Generating Mechanism}
\label{sec:app:syn}
Here, we describe the data-generating mechanism of the synthetic dataset in more detail.
Let $N$, $p$, and $C$ denote the number of independent data points $\left\{\left(\vx_n,\vc_n,y_n\right)\right\}_{n=1}^N$, covariates, and concepts, respectively. We set $N=50$,$000$, $p=1$,$500$, and $C=100$, with a 60\%-20\%-20\% train-validation-test split. The generative process is as follows:
\begin{enumerate}
    \item Randomly sample $\mW\in\mathbb{R}^{C\times 10}$ s.t. $w_{i,j}\sim\mathcal{N}(0,1)$ for $1\leq i\leq C$ and $1\leq j\leq 10$.
    \item Generate a positive definite matrix $\mSigma\in\mathbb{R}^{C\times C}$ s.t. $\mSigma = \mW \mW^T + \mD$. Let $\mD\in\mathbb{R}^{C\times C}$ s.t. $\mD =\boldsymbol{\delta} \mI  $, where $\delta_{i}\sim{\displaystyle {\mathcal {U}}_{[0,1]}}$ for $1\leq i\leq C$.
    \item Randomly sample logits $\mH\in\mathbb{R}^{N\times C}$ s.t. $\boldsymbol{\eta}_{n}\sim\mathcal{N}(\boldsymbol{0},\mSigma)$ for $1\leq n\leq N$.
    \item Let $c_{n,i}=\mathbbm{1}_{\left\{\eta_{n,i}\geq 0\right\}}$ for $1\leq n\leq N$ and $1\leq i\leq C$.
    \item  Let $h:\:\mathbb{R}^{C}\rightarrow\mathbb{R}^p$ be a randomly initialised multilayer perceptron with ReLU nonlinearities.
    \item Let $\vx_n=h\left(\boldsymbol{\eta}_n\right) + \boldsymbol{\epsilon}_n$ s.t. $\boldsymbol{\epsilon}_{n}\sim\mathcal{N}(\boldsymbol{0},\mI)$ for $1\leq n\leq N$.
    \item Let $g:\:\mathbb{R}^C\rightarrow\mathbb{R}$ be a randomly initialized linear perceptron.
    \item Let $y_n=\mathbbm{1}_{\left\{\left(g\left(\vc_n\right)\geq y_{med}\right)\right\}}$ for $1\leq n\leq N$, where $y_{med}$ denotes the median of $g\left(\vc_n\right)$.
\end{enumerate}

\subsection{Natural Image Datasets}
\label{app:nat_ima}
\paragraph{Caltech-UCSD Birds-200-2011} We evaluate on the Caltech-UCSD Birds-200-2011 (CUB)\footnote{\url{https://www.vision.caltech.edu/datasets/cub_200_2011/}, no license available} dataset \citep{wah2011caltech}. It comprises 11,788 photographs from 200 distinct bird species annotated with 312 concepts, such as belly color and pattern. In this manuscript, we follow the original train-test split and revised the proposed dataset in the initial CBM work \citep{Koh2020}. Here, only the 112 most widespread binary attributes are included in the final dataset, and concepts are shared across samples in identical classes. The images were resized to a resolution of 224 × 224 pixels. Finally, following the original proposed augmentations, we applied random horizontal flips, modified the brightness and saturation, and applied normalization during training.

\paragraph{CIFAR-10} CIFAR-10\footnote{\url{https://www.cs.toronto.edu/~kriz/cifar.html}, no license available}~\citep{krizhevsky2009learning} is a natural image benchmark with 60,000 32x32 colour images and 10 classes. We kept the original train-test split, with 50,000 samples in the train set and a balanced total of 6,000 images per class. We generated 143 concept labels as described in Section~\ref{sec:exp_setup} using large language and vision models. At training time, as for CUB, we applied augmentations including modifications to brightness and saturation, random horizontal flips and normalisation. Images were rescaled to a size of 224 × 224 pixels.

\section{Implementation Details}
\label{app:imp_det}
This section provides further implementation details of SCBM and the evaluated baselines. All methods were implemented using PyTorch (v 2.1.1) \citep{ansel2024pytorch}. All models are trained for 150 epochs for the synthetic and 300 epochs for the natural image datasets with the Adam optimizer~\citep{kingma2014adam} with a learning rate of $10^{-4}$ and a batch size of 64. For the independently trained autoregressive model, we split the training epochs into $2/3$ for the concept predictor and $1/3$ for the target predictor. For the methods requiring sampling, the number of Monte Carlo samples is set to $M=100$. We provide an ablation for $M=10$ in Appendix~\ref{app:num_mcmc}. Note that since the predictor head is very simple, the MC sampling of SCBMs is extremely fast and does not influence computational complexity by more than $0.1\%$. For the synthetic tabular data, we use a fully connected neural network as backbone, with 3 non-linear layers, batch normalization, and dropout. For the CUB dataset, we use a pretrained ResNet-18 \citep{he2016deep}, and for the lower-resolution CIFAR-10 a simple convolutional neural network with 2 convolutional layers followed by ReLU, Dropout, and a fully connected layer. For fairness in the comparisons, all baselines have the same model architecture choices and all experiments are performed over $10$ random seeds.

\paragraph{Resource Usage} For the experiments of the main paper, we used a cluster of mostly GeForce RTX 2080s with 2 CPU workers. Over all methods, we estimate an average runtime of 8h per experiment, each running on a single GPU. This amounts to 5 methods $\times$ 3 datasets $\times$ 10 seeds $\times$ 8 hours $=$ 1200 hours. Adding to that, the Ablation Figures required another 40 runs, amounting to a full total of 1520 hours of compute. Please note that we only report the numbers to generate the final results but not the development time, which we roughly estimate to be around 10 times bigger.

\section{Further Experiments}
In this section, we show additional experiments to provide a more in-depth understanding of SCBM's effectiveness. We ablate multiple hyperparameters to provide an understanding of how they influence the model performance, as well as show the performance of our model in other settings.

\begin{figure*}[htbp]
    \centering
    \begin{minipage}[t]{0.44\linewidth} % Changed alignment to top
        \centering % Changed to centering instead of flushright for symmetry
        \includegraphics[height=4.0cm,keepaspectratio]{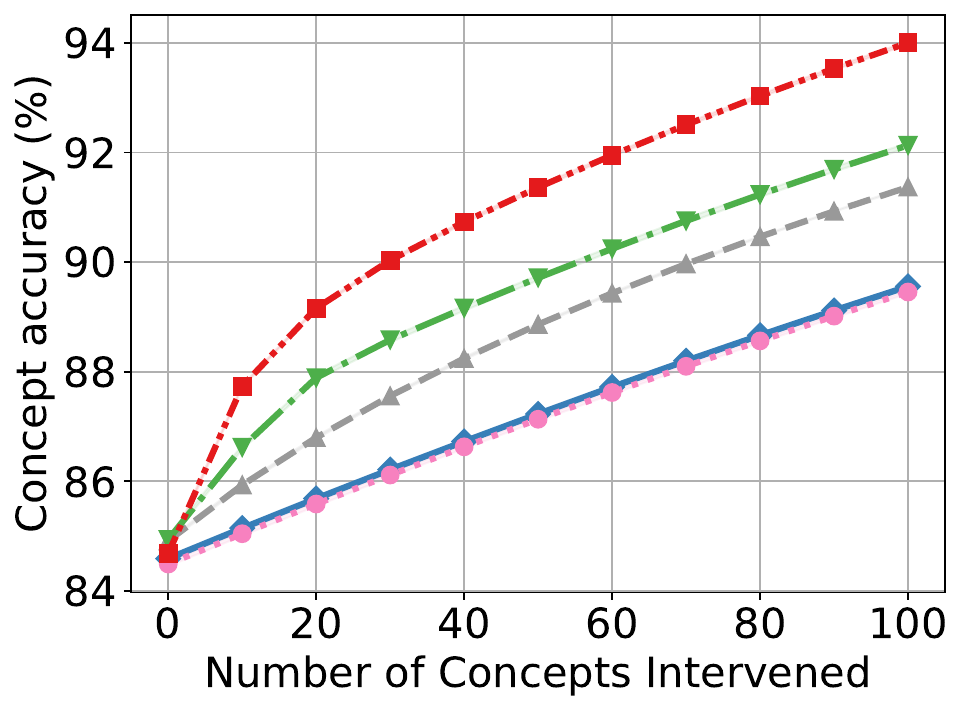}
        \includegraphics[height=4.0cm,keepaspectratio]{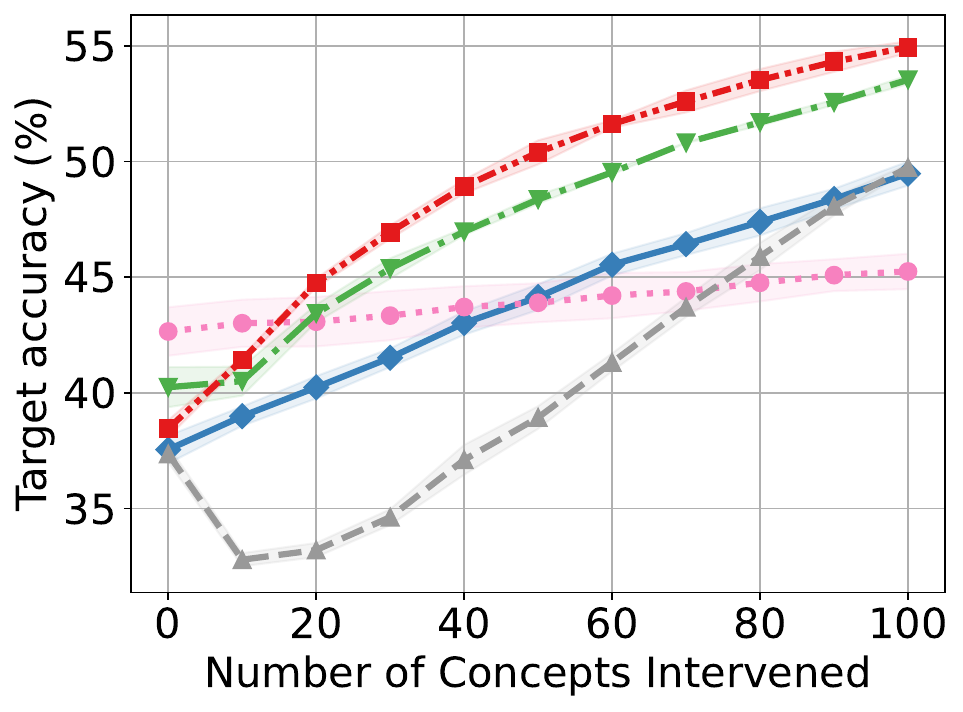}
        \includegraphics[width=1\linewidth]{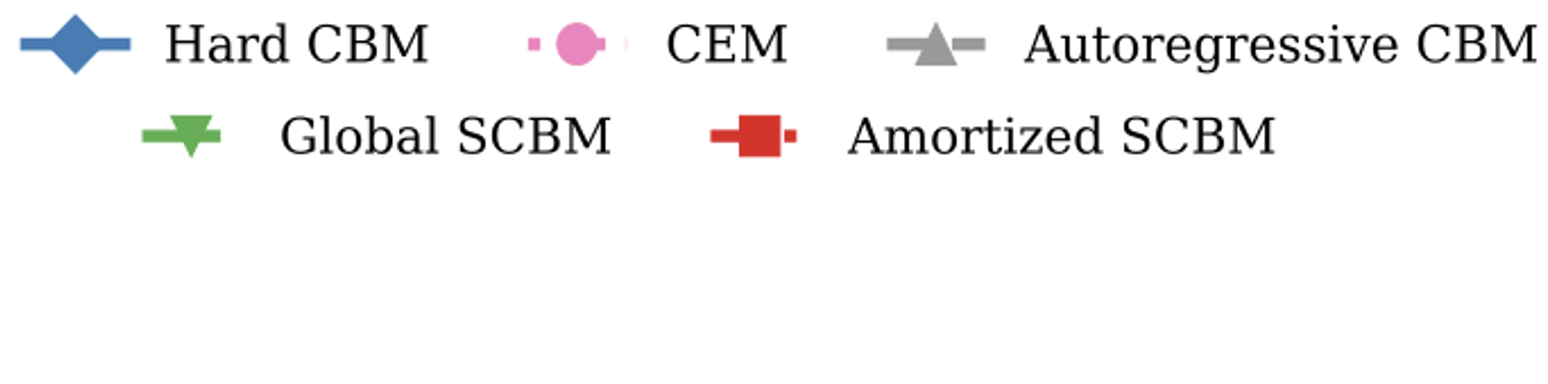}
        \vspace{-0.95cm}
        \caption{Performance after intervening on concepts in the order of highest predicted uncertainty in CIFAR-100 with 892 concepts.
Concept and target accuracy (\%) are shown in the first and second rows, respectively. Results are
reported as averages and standard deviations of model performance across 3 seeds.}
        \label{fig:cifar100}
    \end{minipage}
    \hspace{0.03\linewidth}
    %\textbf{Uncertainty-based Interventions}\par\medskip
    \begin{minipage}[t]{0.46\linewidth} % Changed alignment to top
        \centering % Changed to centering instead of flushright for symmetry
        \includegraphics[height=4.0cm,keepaspectratio]{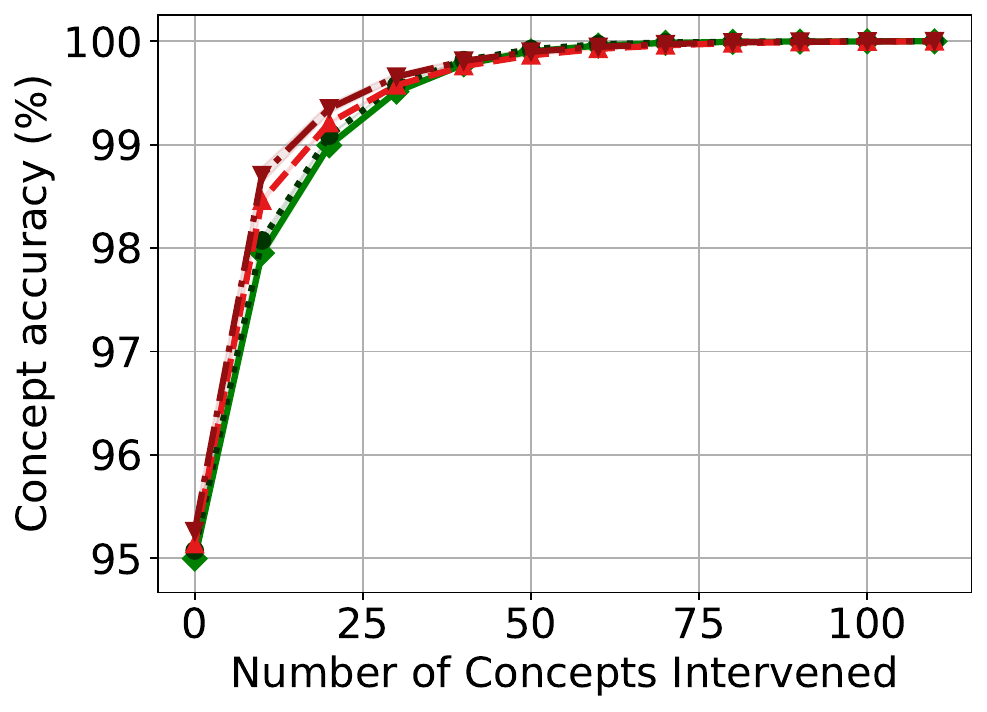}
        \includegraphics[height=4.0cm,keepaspectratio]{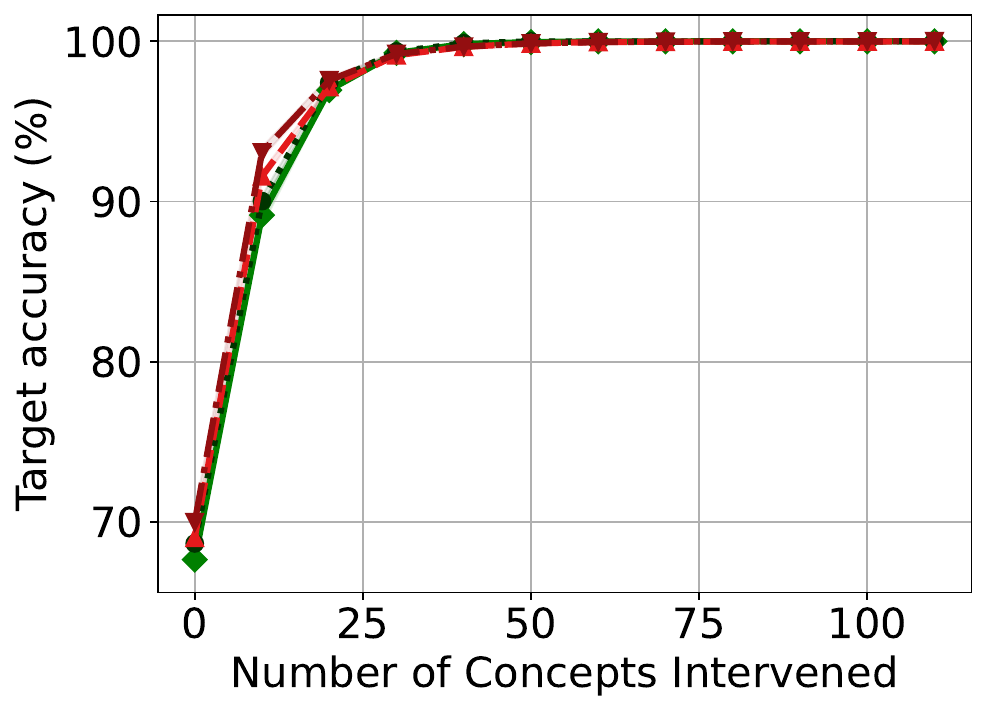}
        \includegraphics[width=1\linewidth]{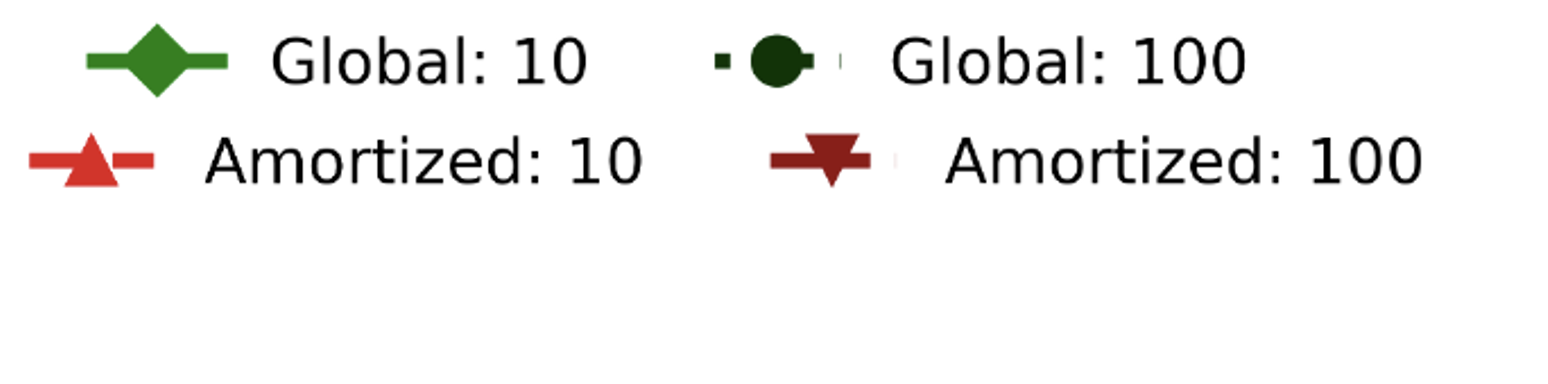}
        \vspace{-1cm}
        \caption{Intervention performance in the order of highest predicted uncertainty in CUB.
Concept and target accuracy (\%) are shown in the first and second rows, respectively. Results are
reported as averages and standard deviations of model performance across 3 seeds.}
        \label{fig:num_mcmc}
    \end{minipage}
\end{figure*}

\subsection{Intervention Performance on CIFAR-100}
\label{app:cifar100}
We present the result on the CIFAR-100 dataset with 892 concepts obtained from~\citet{oikarinen2023label} in Figure~\ref{fig:cifar100} to showcase the scalability of SCBMs. The results underline the efficiency of our method. Notably, the Autoregressive baseline has a negative dip, which is likely due to the independently trained target predictor not being aligned with the concept predictors in this noisy CLIP-annotated scenario. Note that they need to train independently to avoid the sequential MC sampling during training, which would otherwise increase training time significantly. Our jointly trained SCBMs do not have this issue and surpass the baselines. We use the same configuration as for CIFAR-10, with the exception that we set $M=10$ to reduce the memory requirement.

\subsection{Number of Monte Carlo Samples}
\label{app:num_mcmc}
To showcase that SCBMs do not rely on a huge number of Monte Carlo samples, we provide an ablation of $M$ in Figure~\ref{fig:num_mcmc}. It shows that even for $M=10$, SCBMs thrive. Note, however, that since $M$ is not a driving factor of SCBMs computational cost, one can leave it at a high number.

\subsection{Random Intervention Policy}
\label{sec:app:rand}
\begin{figure*}[!htb]
    \centering
    %\textbf{Random Interventions}\par\medskip
    \begin{subfigure}[t]{0.32\linewidth} % Changed alignment to top
        \centering % Changed to centering instead of flushright for symmetry
        \includegraphics[height=3.3cm,keepaspectratio,right]{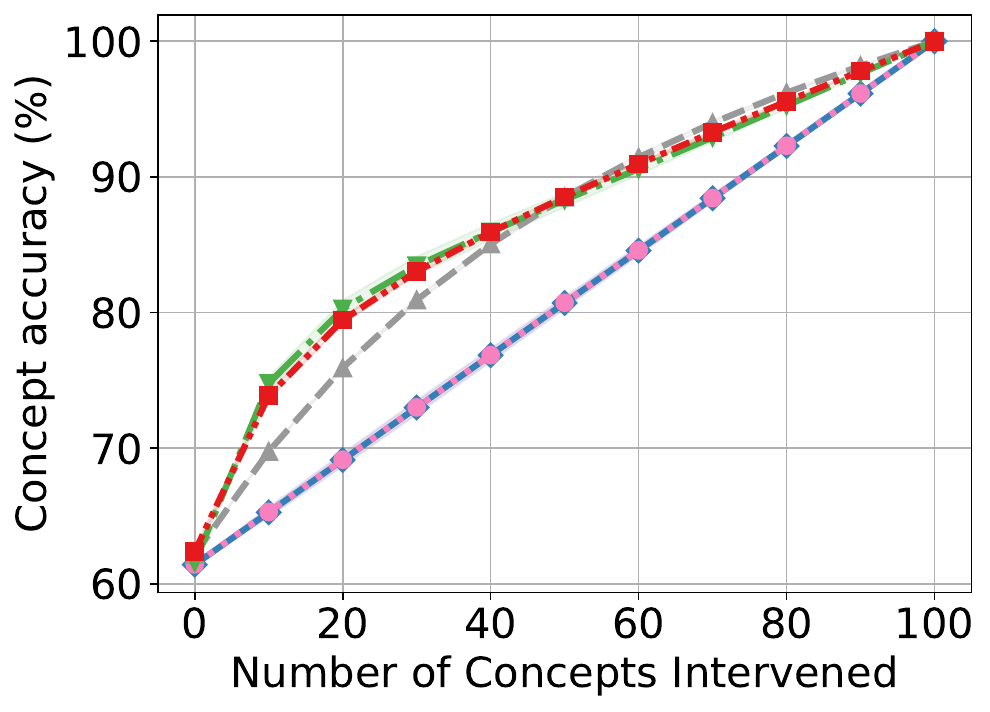}
        \includegraphics[height=3.3cm,keepaspectratio,right]{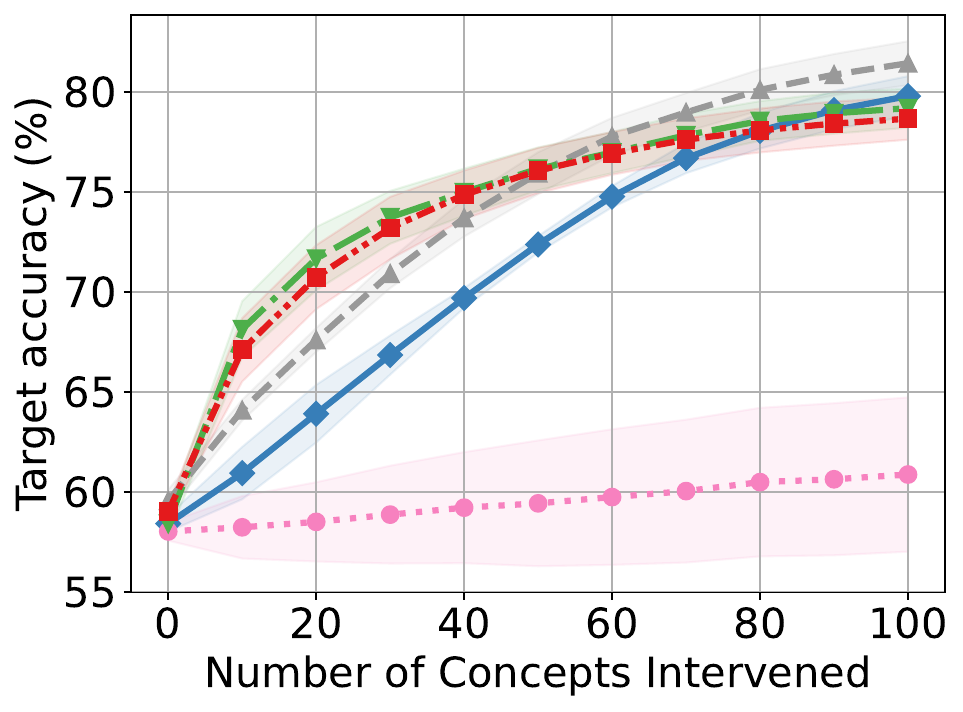}
        \caption{\footnotesize {Synthetic}}
    \end{subfigure}
    \begin{subfigure}[t]{0.32\linewidth} % Changed alignment to top
        \centering % Changed to centering instead of flushright for symmetry
        \includegraphics[height=3.3cm,keepaspectratio,right]{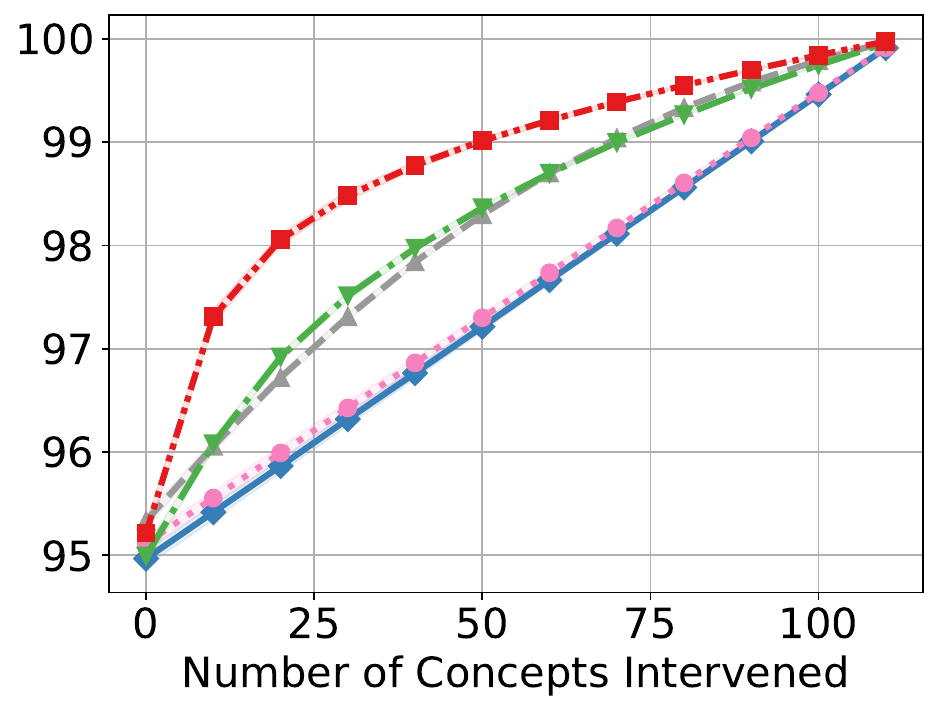}
        \includegraphics[height=3.3cm,keepaspectratio,right]{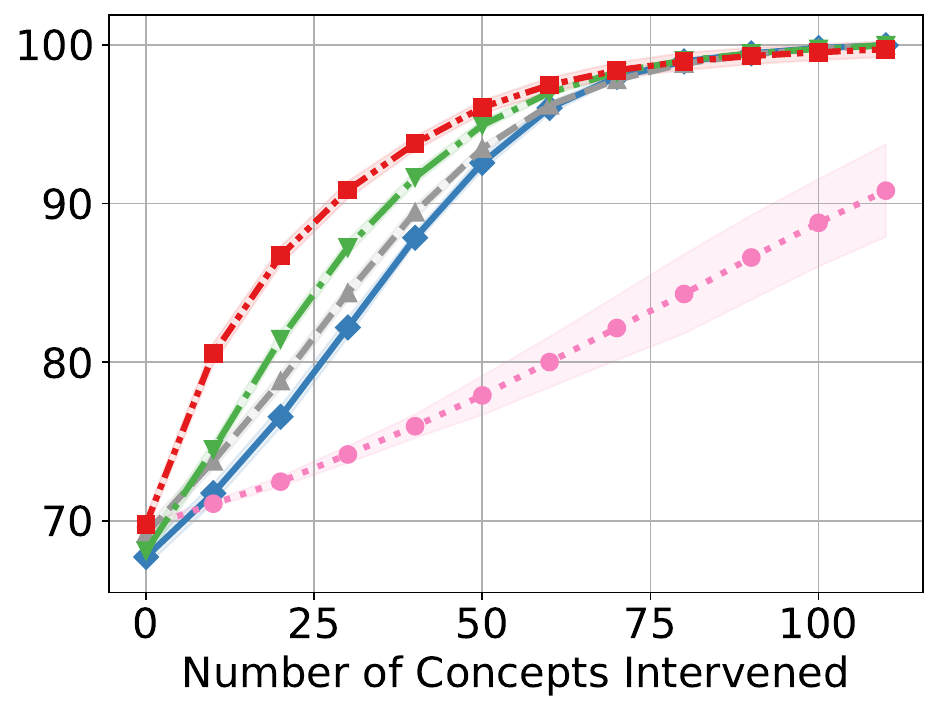}
        \caption{\footnotesize {CUB}}
    \end{subfigure}
    \begin{subfigure}[t]{0.32\linewidth} % Changed alignment to top
        \centering % Changed to centering instead of flushright for symmetry
        \includegraphics[height=3.3cm,keepaspectratio,right]{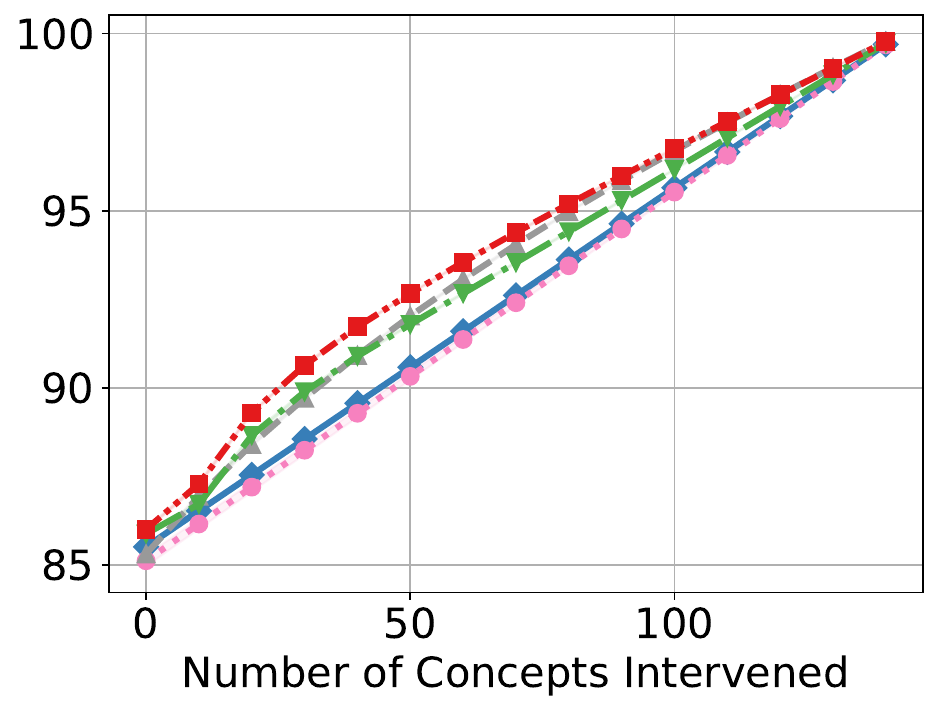}
        \includegraphics[height=3.3cm,keepaspectratio,right]{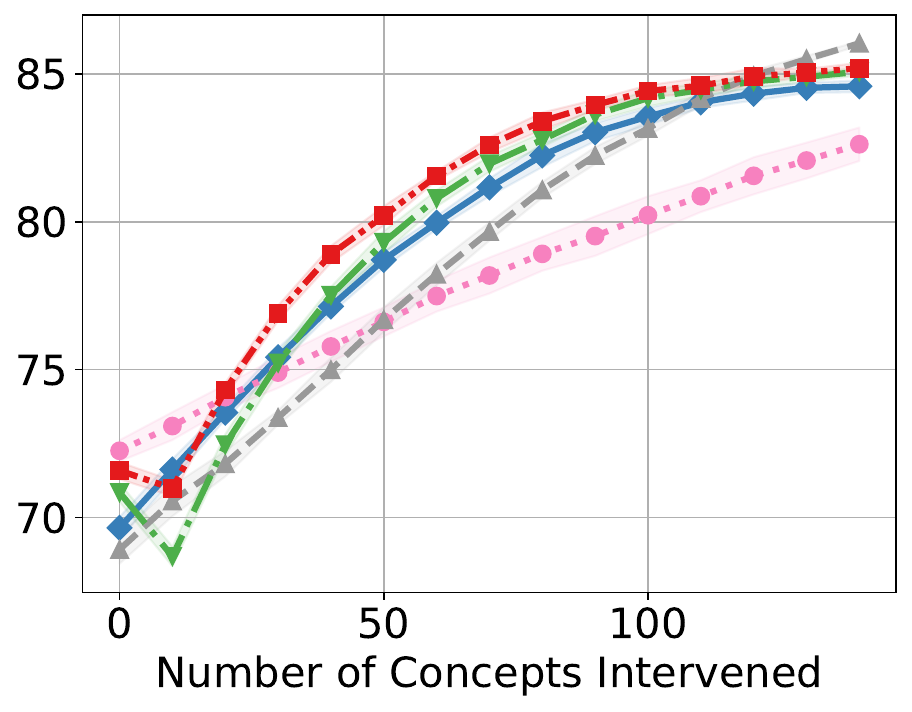}
        \caption{\footnotesize {CIFAR-10}}
    \end{subfigure}
    % \vspace{-0.25cm}
    \includegraphics[width=0.9\linewidth]{Figures/legend_new.pdf}
    \caption{Performance after intervening on concepts in random order. Concept and target accuracy (\%) are shown in the first and second rows, respectively. Results are reported as averages and standard deviations of model performance across ten seeds.}
    \label{inter_plots_random}
\end{figure*}
In Figure~\ref{inter_plots_random}, we present the intervention performance of SCBM and baseline methods. Compared to the uncertainty-based intervention policy of Figure~\ref{inter_plots_uncert}, the intervention curves of all methods are less steep, confirming the usefulness of \citet{Shin2023}'s proposed policy. Following the previous statements, SCBMs still outperform baseline methods with the amortized beating the global variant for real-world datasets. We observe that in CIFAR-10 for the first interventions, an improvement in concept accuracy is not directly reflected in improved target prediction for SCBMs, which is likely due to the low signal-to-noise ratio of the CLIP-inferred concepts.

\subsection{Regularization Strength}
\label{app:reg_strength}
\begin{figure*}[!htb]
    \centering
    %\textbf{Uncertainty-based Interventions}\par\medskip
    \includegraphics[height=3.3cm,keepaspectratio]{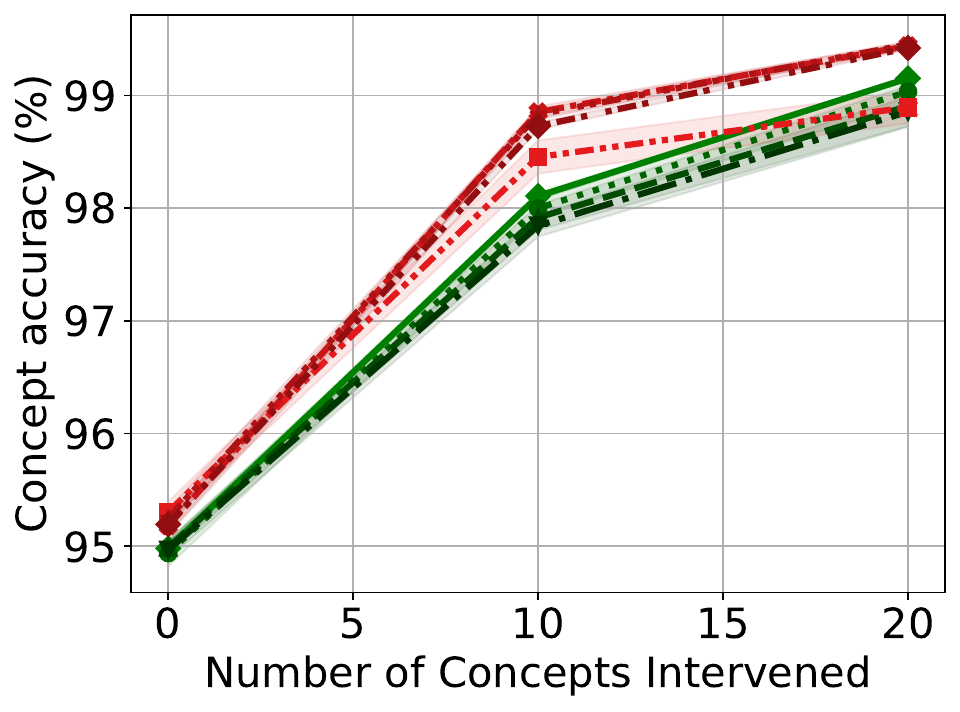}
    \includegraphics[height=3.3cm,keepaspectratio]{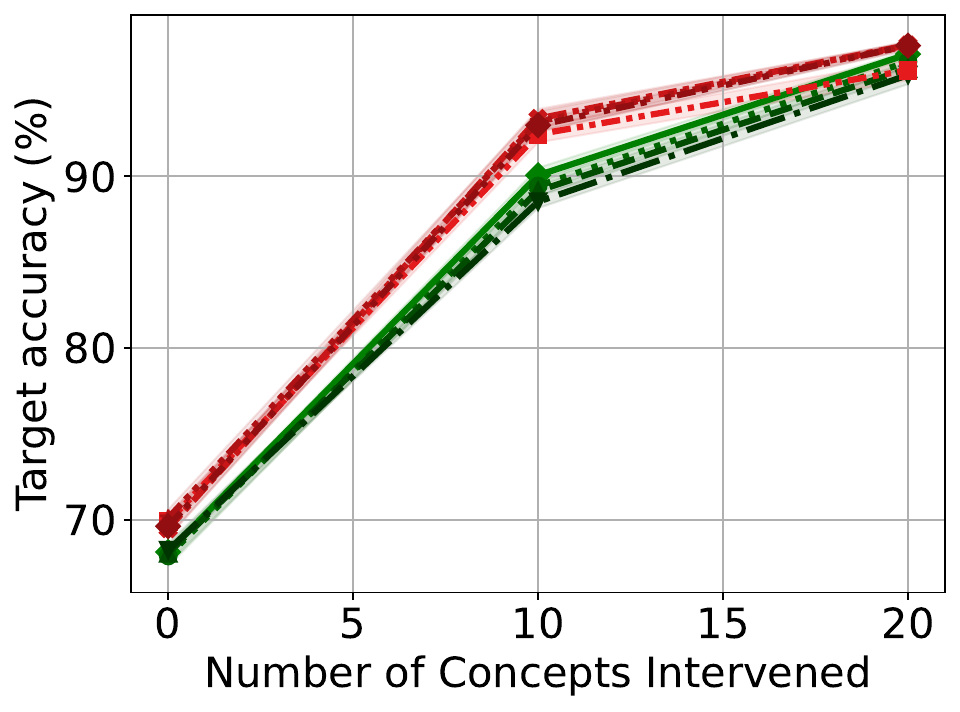}
    \includegraphics[width=0.8\linewidth]{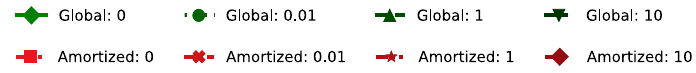}
    \caption{Performance on CUB after intervening on concepts in the order of highest predicted uncertainty with differing regularization strengths. Concept and target accuracy (\%) are shown in the first and second columns, respectively. Results are reported as averages and standard deviations of model performance across five seeds. For each SCBM variant, we choose a darker color, the higher the regularization strength of $\lambda_2$.}
    \label{inter_plots_reg_ablation}
\end{figure*}
In Figure~\ref{inter_plots_reg_ablation}, we analyze the impact of the strength of $\lambda_2$ from Equation~\ref{eq:loss}. Due to environmental considerations, we conducted experiments using only 5 seeds and limited the number of interventions to 20. Our findings indicate that SCBMs are not sensitive to the choice of $\lambda_2$, except that the unregularized amortized variant exhibits slight patterns of overfitting.
\vspace{2cm}
\subsection{Intervention Strategy}
\begin{figure*}[!htb]
    \centering
    \includegraphics[height=3.3cm,keepaspectratio]{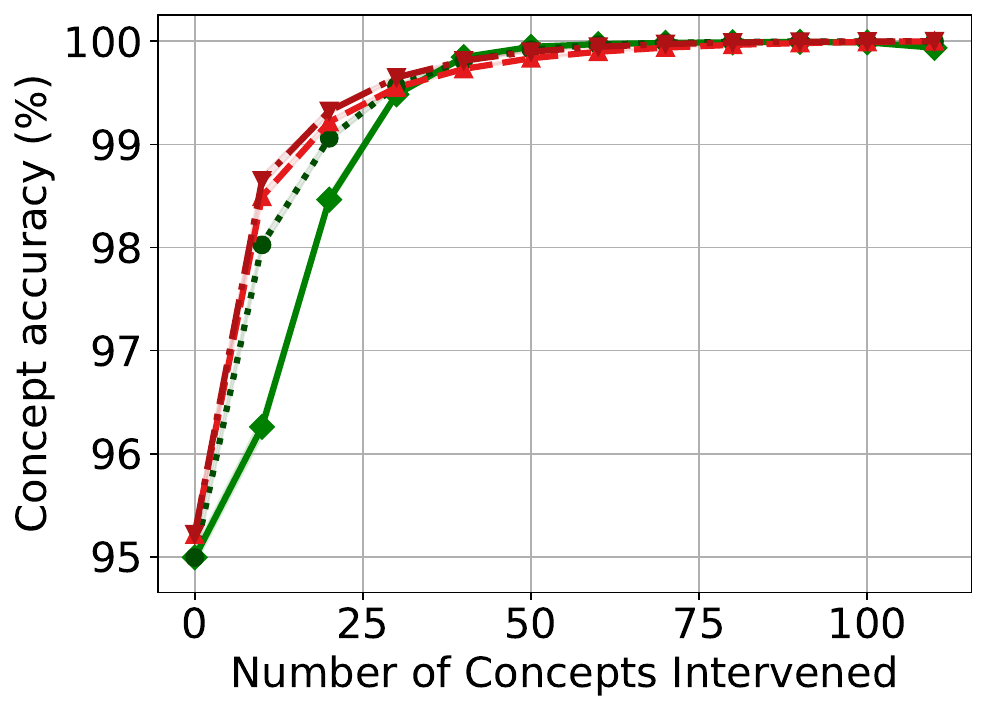}
    \includegraphics[height=3.3cm,keepaspectratio]{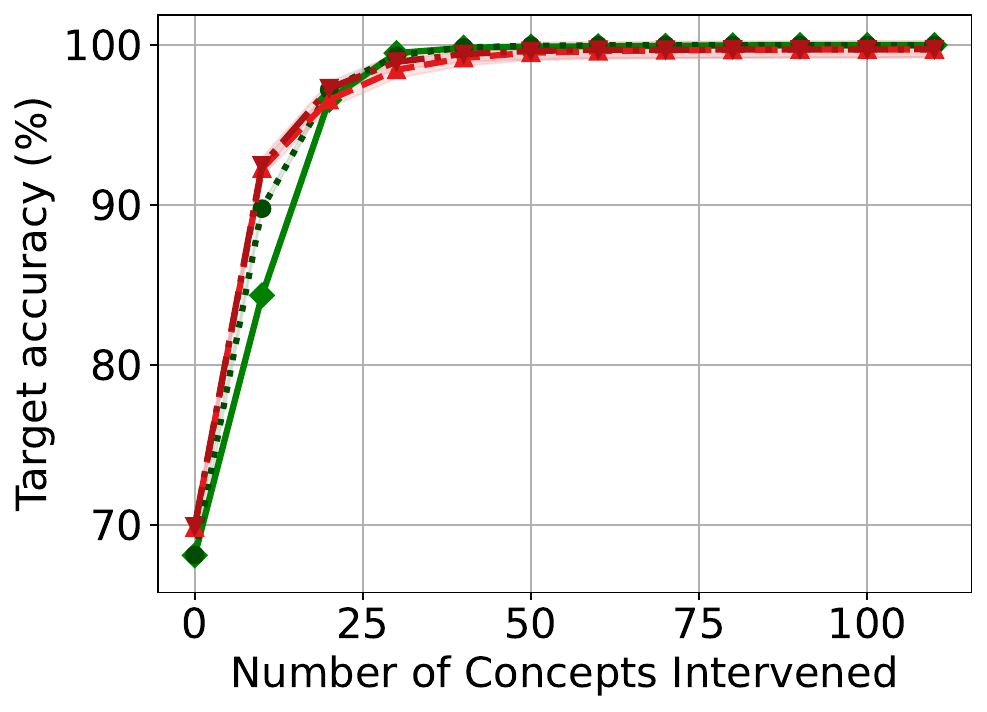}
    \includegraphics[width=0.8\linewidth]{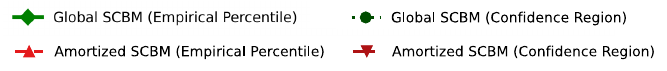}
    \caption{Performance on CUB after intervening on concepts in the order of highest predicted uncertainty, comparing the proposed intervention strategy to \citet{Koh2020}'s intervention of setting the logits to the 5th or 95th empirical percentile of the training distribution. Concept and target accuracy (\%) are shown in the first and second columns, respectively. Results are reported as averages and standard deviations of model performance across five seeds.}
    \label{inter_plots_interv_strats}
\end{figure*}

In Figure~\ref{inter_plots_interv_strats}, we analyze the effect of the intervention strategy. Our findings indicate that while SCBMs are still effective with the proposed strategy from \citet{Koh2020}, that sets the logits to the 5th (if $c_i = 0$) or 95th (if $c_i = 1$) percentile of the training distribution, our proposed strategy based on the confidence region results in stronger intervenability.

\label{app:interv_strat}

\subsection{Confidence Region Level}
\label{app:conf_region_level}
\begin{figure*}[!htb]
    \centering
    \includegraphics[height=3.3cm,keepaspectratio]{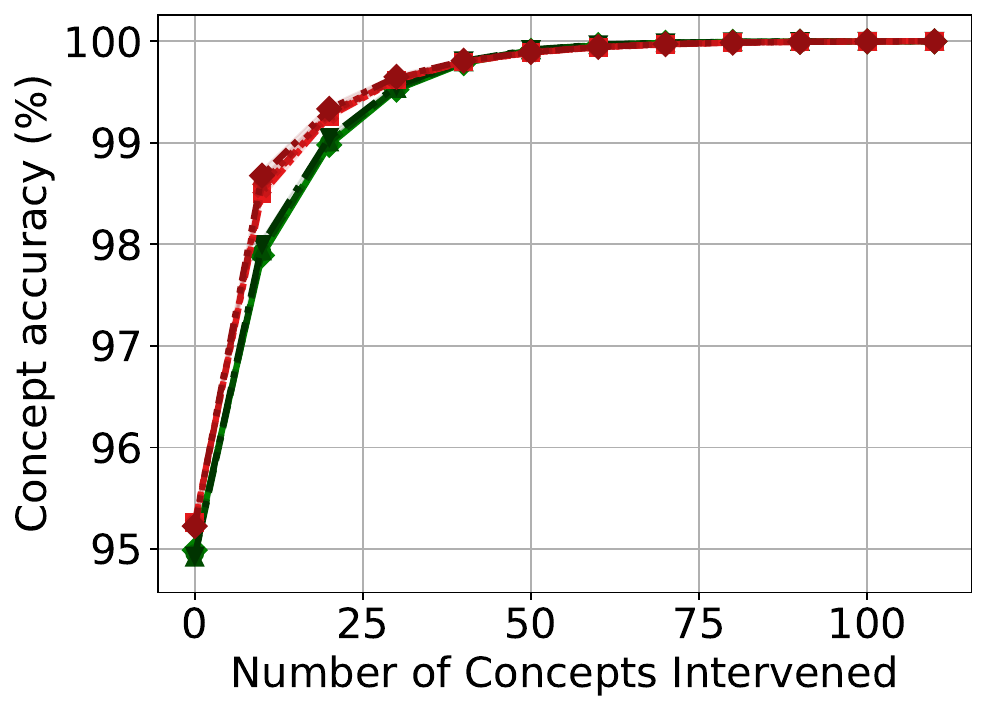}
    \includegraphics[height=3.3cm,keepaspectratio]{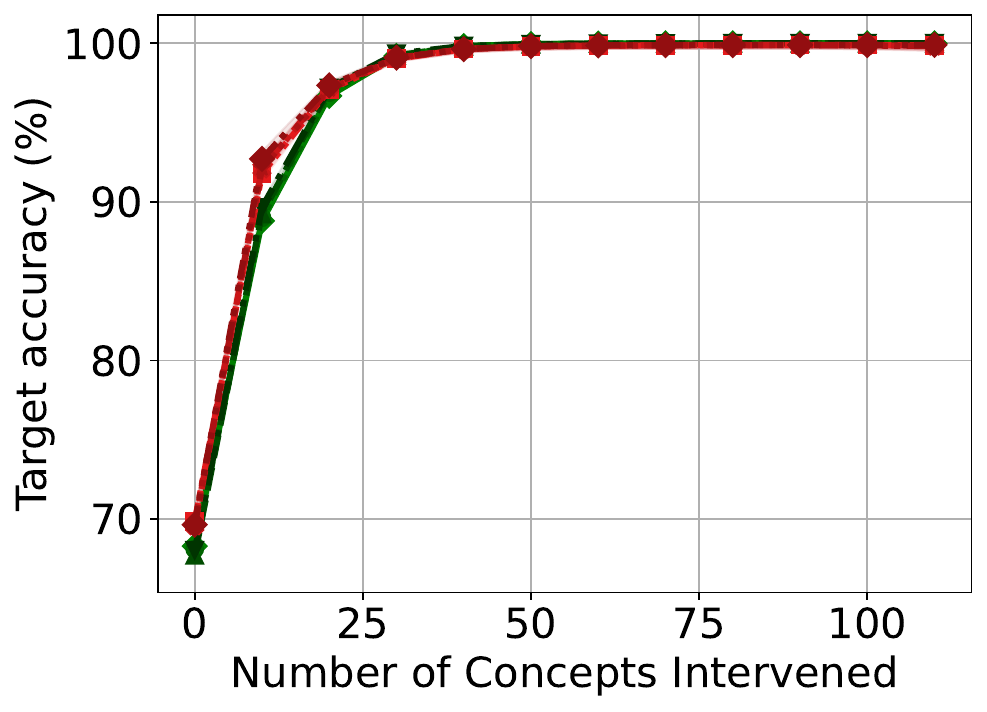}
    \includegraphics[width=0.8\linewidth]{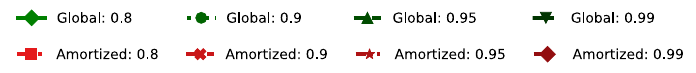}
    \caption{Performance on CUB after intervening on concepts in the order of highest predicted uncertainty with differing levels $1-\alpha$ of the confidence region. Concept and target accuracy (\%) are shown in the first and second columns, respectively. Results are reported as averages and standard deviations of model performance across three seeds.}
    \label{inter_plots_level}
\end{figure*}
In Figure~\ref{inter_plots_level}, we analyze the effect of the level $1-\alpha$ of the likelihood-based confidence region. Our findings indicate that the SCBMs are not sensitive to the choice of $1-\alpha$, with higher levels being slightly better in performance.
%TODO: Check that all figures and section are mentioned in main text

\subsection{Jaccard Index}
\label{app:jaccard}
\citet{panousis2024coarse} propose to interpret the interpretation capacity of concepts with the Jaccard Index~\citep{jaccard1901etude}. As such, in Table~\ref{tab:test_performance_jac}, we extend Table~\ref{tab:test_performance} with this metric. It is evident that the interpretation does not change, indicating that the performance is robust to the choice of evaluation metric.

\begin{table}[htbp]
\caption{Test-set performance before interventions. Results are averaged across ten seeds.}
\label{tab:test_performance_jac}
\centering
\begin{tabular}{llccc}
\toprule
Dataset & Method & Concept Accuracy & Concept Jaccard & Target Accuracy  \\
\midrule
 & Hard CBM & 61.42 \small{$\pm$ 0.07} & 43.80 \small{$\pm$ 1.32} & 58.38 \small{$\pm$ 0.39} \\
 & CEM & 61.42 \small{$\pm$ 0.12} & 44.84 \small{$\pm$ 1.36} & 58.01 \small{$\pm$ 0.49} \\
Synthetic & Autoregressive CBM & \underline{62.17} \small{$\pm$ 0.11} & \underline{45.30} \small{$\pm$ 1.29} & \textbf{59.60} \small{$\pm$ 0.62} \\
 & Global SCBM & 61.57 \small{$\pm$ 0.05} & 44.53 \small{$\pm$ 1.02} & 58.39 \small{$\pm$ 0.53} \\
 & Amortized SCBM & \textbf{62.41} \small{$\pm$ 0.20} & \textbf{45.85} \small{$\pm$ 1.45} & \underline{58.96} \small{$\pm$ 0.38} \\
\midrule
 & Hard CBM & 94.97 \small{$\pm$ 0.07} & 77.22 \small{$\pm$ 0.33} & 67.72 \small{$\pm$ 0.57} \\
 & CEM & 95.12 \small{$\pm$ 0.07} & 78.20 \small{$\pm$ 0.28} & \underline{69.60} \small{$\pm$ 0.30} \\
CUB & Autoregressive CBM & \textbf{95.33} \small{$\pm$ 0.07} & \textbf{79.21} \small{$\pm$ 0.21} & 69.24 \small{$\pm$ 0.44} \\
 & Global SCBM & 94.99 \small{$\pm$ 0.09} & 76.83 \small{$\pm$ 0.47} & 68.19 \small{$\pm$ 0.63} \\
 & Amortized SCBM & \underline{95.22} \small{$\pm$ 0.09} & \underline{78.29} \small{$\pm$ 0.28} & \textbf{69.87} \small{$\pm$ 0.56} \\
\midrule
 & Hard CBM & 85.51 \small{$\pm$ 0.04} & 81.54 \small{$\pm$ 0.08} & 69.73 \small{$\pm$ 0.29} \\
 & CEM & 85.12 \small{$\pm$ 0.14} & 81.06 \small{$\pm$ 0.21} & \textbf{72.24} \small{$\pm$ 0.33} \\
CIFAR-10 & Autoregressive CBM & 85.31 \small{$\pm$ 0.06} & 81.31 \small{$\pm$ 0.10} & 68.88 \small{$\pm$ 0.47} \\
 & Global SCBM & \underline{85.86} \small{$\pm$ 0.04} & \underline{81.81} \small{$\pm$ 0.19} & 70.74 \small{$\pm$ 0.29} \\
 & Amortized SCBM & \textbf{86.00} \small{$\pm$ 0.03} & \textbf{81.97} \small{$\pm$ 0.20} & \underline{71.66} \small{$\pm$ 0.25} \\
\bottomrule
\end{tabular}
\end{table}

%% file: main.bbl
\begin{thebibliography}{}

\bibitem [\protect \citeauthoryear {%
Ansel%
\ \protect \BOthers {.}}{%
Ansel%
\ \protect \BOthers {.}}{%
{\protect \APACyear {2024}}%
}]{%
ansel2024pytorch}
\APACinsertmetastar {%
ansel2024pytorch}%
\begin{APACrefauthors}%
Ansel, J.%
, Yang, E.%
, He, H.%
, Gimelshein, N.%
, Jain, A.%
, Voznesensky, M.%
\BDBL {}others%
\end{APACrefauthors}%
\unskip\
\newblock
\APACrefYearMonthDay{2024}{}{}.
\newblock
{\BBOQ}\APACrefatitle {PyTorch 2: Faster Machine Learning Through Dynamic Python Bytecode Transformation and Graph Compilation} {Pytorch 2: Faster machine learning through dynamic python bytecode transformation and graph compilation}.{\BBCQ}
\newblock
\BIn{} \APACrefbtitle {Proceedings of the 29th ACM International Conference on Architectural Support for Programming Languages and Operating Systems, Volume 2} {Proceedings of the 29th acm international conference on architectural support for programming languages and operating systems, volume 2}\ (\BPGS\ 929--947).
\PrintBackRefs{\CurrentBib}

\bibitem [\protect \citeauthoryear {%
Brier%
}{%
Brier%
}{%
{\protect \APACyear {1950}}%
}]{%
brier1950verification}
\APACinsertmetastar {%
brier1950verification}%
\begin{APACrefauthors}%
Brier, G\BPBI W.%
\end{APACrefauthors}%
\unskip\
\newblock
\APACrefYearMonthDay{1950}{}{}.
\newblock
{\BBOQ}\APACrefatitle {Verification of forecasts expressed in terms of probability} {Verification of forecasts expressed in terms of probability}.{\BBCQ}
\newblock
\APACjournalVolNumPages{Monthly weather review}{78}{1}{1--3}.
\newblock
\begin{APACrefURL} \url{https://doi.org/10.1175/1520-0493(1950)078<0001:VOFEIT>2.0.CO;2} \end{APACrefURL}
\PrintBackRefs{\CurrentBib}

\bibitem [\protect \citeauthoryear {%
Brown%
\ \protect \BOthers {.}}{%
Brown%
\ \protect \BOthers {.}}{%
{\protect \APACyear {2020}}%
}]{%
brown2020language}
\APACinsertmetastar {%
brown2020language}%
\begin{APACrefauthors}%
Brown, T.%
, Mann, B.%
, Ryder, N.%
, Subbiah, M.%
, Kaplan, J\BPBI D.%
, Dhariwal, P.%
\BDBL {}others%
\end{APACrefauthors}%
\unskip\
\newblock
\APACrefYearMonthDay{2020}{}{}.
\newblock
{\BBOQ}\APACrefatitle {Language models are few-shot learners} {Language models are few-shot learners}.{\BBCQ}
\newblock
\APACjournalVolNumPages{Advances in neural information processing systems}{33}{}{1877--1901}.
\PrintBackRefs{\CurrentBib}

\bibitem [\protect \citeauthoryear {%
Chauhan%
, Tiwari%
, Freyberg%
, Shenoy%
\BCBL {}\ \BBA {} Dvijotham%
}{%
Chauhan%
\ \protect \BOthers {.}}{%
{\protect \APACyear {2023}}%
}]{%
chauhan2023interactive}
\APACinsertmetastar {%
chauhan2023interactive}%
\begin{APACrefauthors}%
Chauhan, K.%
, Tiwari, R.%
, Freyberg, J.%
, Shenoy, P.%
\BCBL {}\ \BBA {} Dvijotham, K.%
\end{APACrefauthors}%
\unskip\
\newblock
\APACrefYearMonthDay{2023}{}{}.
\newblock
{\BBOQ}\APACrefatitle {Interactive concept bottleneck models} {Interactive concept bottleneck models}.{\BBCQ}
\newblock
\BIn{} \APACrefbtitle {Proceedings of the AAAI Conference on Artificial Intelligence} {Proceedings of the aaai conference on artificial intelligence}\ (\BVOL~37, \BPGS\ 5948--5955).
\PrintBackRefs{\CurrentBib}

\bibitem [\protect \citeauthoryear {%
Collins%
\ \protect \BOthers {.}}{%
Collins%
\ \protect \BOthers {.}}{%
{\protect \APACyear {2023}}%
}]{%
Collins2023}
\APACinsertmetastar {%
Collins2023}%
\begin{APACrefauthors}%
Collins, K\BPBI M.%
, Barker, M.%
, Zarlenga, M\BPBI E.%
, Raman, N.%
, Bhatt, U.%
, Jamnik, M.%
\BDBL {}Dvijotham, K.%
\end{APACrefauthors}%
\unskip\
\newblock
\APACrefYearMonthDay{2023}{}{}.
\newblock
{\BBOQ}\APACrefatitle {Human Uncertainty in Concept-Based {AI} Systems} {Human uncertainty in concept-based {AI} systems}.{\BBCQ}
\newblock
\BIn{} F.~Rossi, S.~Das, J.~Davis, K.~Firth{-}Butterfield\BCBL {}\ \BBA {} A.~John\ (\BEDS), \APACrefbtitle {Proceedings of the 2023 {AAAI/ACM} Conference on AI, Ethics, and Society, {AIES} 2023, Montr{\'{e}}al, QC, Canada, August 8-10, 2023} {Proceedings of the 2023 {AAAI/ACM} conference on ai, ethics, and society, {AIES} 2023, montr{\'{e}}al, qc, canada, august 8-10, 2023}\ (\BPGS\ 869--889).
\newblock
\APACaddressPublisher{}{{ACM}}.
\PrintBackRefs{\CurrentBib}

\bibitem [\protect \citeauthoryear {%
{Doshi-Velez}%
\ \BBA {} Kim%
}{%
{Doshi-Velez}%
\ \BBA {} Kim%
}{%
{\protect \APACyear {2017}}%
}]{%
doshiRigorousScienceInterpretable2017}
\APACinsertmetastar {%
doshiRigorousScienceInterpretable2017}%
\begin{APACrefauthors}%
{Doshi-Velez}, F.%
\BCBT {}\ \BBA {} Kim, B.%
\end{APACrefauthors}%
\unskip\
\newblock
\APACrefYearMonthDay{2017}{{\APACmonth{03}}}{}.
\newblock
\APACrefbtitle {Towards {{A Rigorous Science}} of {{Interpretable Machine Learning}}} {Towards {{A Rigorous Science}} of {{Interpretable Machine Learning}}}\ (\BNUM\ arXiv:1702.08608).
\newblock
\APACaddressPublisher{}{{arXiv}}.
\newblock
\begin{APACrefDOI} \doi{10.48550/arXiv.1702.08608} \end{APACrefDOI}
\PrintBackRefs{\CurrentBib}

\bibitem [\protect \citeauthoryear {%
Espinosa~Zarlenga%
\ \protect \BOthers {.}}{%
Espinosa~Zarlenga%
\ \protect \BOthers {.}}{%
{\protect \APACyear {2022}}%
}]{%
espinosa2022concept}
\APACinsertmetastar {%
espinosa2022concept}%
\begin{APACrefauthors}%
Espinosa~Zarlenga, M.%
, Barbiero, P.%
, Ciravegna, G.%
, Marra, G.%
, Giannini, F.%
, Diligenti, M.%
\BDBL {}others%
\end{APACrefauthors}%
\unskip\
\newblock
\APACrefYearMonthDay{2022}{}{}.
\newblock
{\BBOQ}\APACrefatitle {Concept embedding models: Beyond the accuracy-explainability trade-off} {Concept embedding models: Beyond the accuracy-explainability trade-off}.{\BBCQ}
\newblock
\BIn{} \APACrefbtitle {Advances in Neural Information Processing Systems} {Advances in neural information processing systems}\ (\BVOL~35, \BPGS\ 21400--21413).
\PrintBackRefs{\CurrentBib}

\bibitem [\protect \citeauthoryear {%
Espinosa~Zarlenga%
\ \protect \BOthers {.}}{%
Espinosa~Zarlenga%
\ \protect \BOthers {.}}{%
{\protect \APACyear {2024}}%
}]{%
espinosa2024learning}
\APACinsertmetastar {%
espinosa2024learning}%
\begin{APACrefauthors}%
Espinosa~Zarlenga, M.%
, Collins, K.%
, Dvijotham, K.%
, Weller, A.%
, Shams, Z.%
\BCBL {}\ \BBA {} Jamnik, M.%
\end{APACrefauthors}%
\unskip\
\newblock
\APACrefYearMonthDay{2024}{}{}.
\newblock
{\BBOQ}\APACrefatitle {Learning to Receive Help: Intervention-Aware Concept Embedding Models} {Learning to receive help: Intervention-aware concept embedding models}.{\BBCQ}
\newblock
\APACjournalVolNumPages{Advances in Neural Information Processing Systems}{36}{}{}.
\PrintBackRefs{\CurrentBib}

\bibitem [\protect \citeauthoryear {%
Friedman%
, Hastie%
\BCBL {}\ \BBA {} Tibshirani%
}{%
Friedman%
\ \protect \BOthers {.}}{%
{\protect \APACyear {2008}}%
}]{%
friedman2008sparse}
\APACinsertmetastar {%
friedman2008sparse}%
\begin{APACrefauthors}%
Friedman, J.%
, Hastie, T.%
\BCBL {}\ \BBA {} Tibshirani, R.%
\end{APACrefauthors}%
\unskip\
\newblock
\APACrefYearMonthDay{2008}{}{}.
\newblock
{\BBOQ}\APACrefatitle {Sparse inverse covariance estimation with the graphical lasso} {Sparse inverse covariance estimation with the graphical lasso}.{\BBCQ}
\newblock
\APACjournalVolNumPages{Biostatistics}{9}{3}{432--441}.
\PrintBackRefs{\CurrentBib}

\bibitem [\protect \citeauthoryear {%
Havasi%
, Parbhoo%
\BCBL {}\ \BBA {} Doshi-Velez%
}{%
Havasi%
\ \protect \BOthers {.}}{%
{\protect \APACyear {2022}}%
}]{%
Havasi2022}
\APACinsertmetastar {%
Havasi2022}%
\begin{APACrefauthors}%
Havasi, M.%
, Parbhoo, S.%
\BCBL {}\ \BBA {} Doshi-Velez, F.%
\end{APACrefauthors}%
\unskip\
\newblock
\APACrefYearMonthDay{2022}{}{}.
\newblock
{\BBOQ}\APACrefatitle {Addressing Leakage in Concept Bottleneck Models} {Addressing leakage in concept bottleneck models}.{\BBCQ}
\newblock
\BIn{} A\BPBI H.~Oh, A.~Agarwal, D.~Belgrave\BCBL {}\ \BBA {} K.~Cho\ (\BEDS), \APACrefbtitle {Advances in Neural Information Processing Systems.} {Advances in neural information processing systems.}
\newblock
\begin{APACrefURL} \url{https://openreview.net/forum?id=tglniD_fn9} \end{APACrefURL}
\PrintBackRefs{\CurrentBib}

\bibitem [\protect \citeauthoryear {%
He%
, Zhang%
, Ren%
\BCBL {}\ \BBA {} Sun%
}{%
He%
\ \protect \BOthers {.}}{%
{\protect \APACyear {2016}}%
}]{%
he2016deep}
\APACinsertmetastar {%
he2016deep}%
\begin{APACrefauthors}%
He, K.%
, Zhang, X.%
, Ren, S.%
\BCBL {}\ \BBA {} Sun, J.%
\end{APACrefauthors}%
\unskip\
\newblock
\APACrefYearMonthDay{2016}{}{}.
\newblock
{\BBOQ}\APACrefatitle {Deep residual learning for image recognition} {Deep residual learning for image recognition}.{\BBCQ}
\newblock
\BIn{} \APACrefbtitle {Proceedings of the IEEE conference on computer vision and pattern recognition} {Proceedings of the ieee conference on computer vision and pattern recognition}\ (\BPGS\ 770--778).
\PrintBackRefs{\CurrentBib}

\bibitem [\protect \citeauthoryear {%
Heidemann%
, Monnet%
\BCBL {}\ \BBA {} Roscher%
}{%
Heidemann%
\ \protect \BOthers {.}}{%
{\protect \APACyear {2023}}%
}]{%
heidemann2023concept}
\APACinsertmetastar {%
heidemann2023concept}%
\begin{APACrefauthors}%
Heidemann, L.%
, Monnet, M.%
\BCBL {}\ \BBA {} Roscher, K.%
\end{APACrefauthors}%
\unskip\
\newblock
\APACrefYearMonthDay{2023}{}{}.
\newblock
{\BBOQ}\APACrefatitle {Concept correlation and its effects on concept-based models} {Concept correlation and its effects on concept-based models}.{\BBCQ}
\newblock
\BIn{} \APACrefbtitle {Proceedings of the IEEE/CVF Winter Conference on Applications of Computer Vision} {Proceedings of the ieee/cvf winter conference on applications of computer vision}\ (\BPGS\ 4780--4788).
\PrintBackRefs{\CurrentBib}

\bibitem [\protect \citeauthoryear {%
Jaccard%
}{%
Jaccard%
}{%
{\protect \APACyear {1901}}%
}]{%
jaccard1901etude}
\APACinsertmetastar {%
jaccard1901etude}%
\begin{APACrefauthors}%
Jaccard, P.%
\end{APACrefauthors}%
\unskip\
\newblock
\APACrefYearMonthDay{1901}{}{}.
\newblock
{\BBOQ}\APACrefatitle {{\'E}tude comparative de la distribution florale dans une portion des Alpes et des Jura} {{\'E}tude comparative de la distribution florale dans une portion des alpes et des jura}.{\BBCQ}
\newblock
\APACjournalVolNumPages{Bull Soc Vaudoise Sci Nat}{37}{}{547--579}.
\PrintBackRefs{\CurrentBib}

\bibitem [\protect \citeauthoryear {%
Jang%
, Gu%
\BCBL {}\ \BBA {} Poole%
}{%
Jang%
\ \protect \BOthers {.}}{%
{\protect \APACyear {2017}}%
}]{%
Gumbel}
\APACinsertmetastar {%
Gumbel}%
\begin{APACrefauthors}%
Jang, E.%
, Gu, S.%
\BCBL {}\ \BBA {} Poole, B.%
\end{APACrefauthors}%
\unskip\
\newblock
\APACrefYearMonthDay{2017}{}{}.
\newblock
{\BBOQ}\APACrefatitle {Categorical Reparameterization with Gumbel-Softmax} {Categorical reparameterization with gumbel-softmax}.{\BBCQ}
\newblock
\BIn{} \APACrefbtitle {5th International Conference on Learning Representations, {ICLR} 2017, Toulon, France, April 24-26, 2017, Conference Track Proceedings.} {5th international conference on learning representations, {ICLR} 2017, toulon, france, april 24-26, 2017, conference track proceedings.}
\newblock
\APACaddressPublisher{}{OpenReview.net}.
\newblock
\begin{APACrefURL} \url{https://openreview.net/forum?id=rkE3y85ee} \end{APACrefURL}
\PrintBackRefs{\CurrentBib}

\bibitem [\protect \citeauthoryear {%
B.~Kim%
\ \protect \BOthers {.}}{%
B.~Kim%
\ \protect \BOthers {.}}{%
{\protect \APACyear {2018}}%
}]{%
Kim2018}
\APACinsertmetastar {%
Kim2018}%
\begin{APACrefauthors}%
Kim, B.%
, Wattenberg, M.%
, Gilmer, J.%
, Cai, C.%
, Wexler, J.%
, Viegas, F.%
\BCBL {}\ \BBA {} Sayres, R.%
\end{APACrefauthors}%
\unskip\
\newblock
\APACrefYearMonthDay{2018}{}{}.
\newblock
{\BBOQ}\APACrefatitle {Interpretability Beyond Feature Attribution: Quantitative Testing with Concept Activation Vectors ({TCAV})} {Interpretability beyond feature attribution: Quantitative testing with concept activation vectors ({TCAV})}.{\BBCQ}
\newblock
\BIn{} J.~Dy\ \BBA {} A.~Krause\ (\BEDS), \APACrefbtitle {Proceedings of the 35th International Conference on Machine Learning} {Proceedings of the 35th international conference on machine learning}\ (\BVOL~80, \BPGS\ 2668--2677).
\newblock
\APACaddressPublisher{}{PMLR}.
\newblock
\begin{APACrefURL} \url{https://proceedings.mlr.press/v80/kim18d.html} \end{APACrefURL}
\PrintBackRefs{\CurrentBib}

\bibitem [\protect \citeauthoryear {%
E.~Kim%
, Jung%
, Park%
, Kim%
\BCBL {}\ \BBA {} Yoon%
}{%
E.~Kim%
\ \protect \BOthers {.}}{%
{\protect \APACyear {2023}}%
}]{%
kim2023probabilistic}
\APACinsertmetastar {%
kim2023probabilistic}%
\begin{APACrefauthors}%
Kim, E.%
, Jung, D.%
, Park, S.%
, Kim, S.%
\BCBL {}\ \BBA {} Yoon, S.%
\end{APACrefauthors}%
\unskip\
\newblock
\APACrefYearMonthDay{2023}{}{}.
\newblock
{\BBOQ}\APACrefatitle {Probabilistic Concept Bottleneck Models} {Probabilistic concept bottleneck models}.{\BBCQ}
\newblock
\BIn{} A.~Krause, E.~Brunskill, K.~Cho, B.~Engelhardt, S.~Sabato\BCBL {}\ \BBA {} J.~Scarlett\ (\BEDS), \APACrefbtitle {Proceedings of the 40th International Conference on Machine Learning} {Proceedings of the 40th international conference on machine learning}\ (\BVOL~202, \BPGS\ 16521--16540).
\newblock
\APACaddressPublisher{}{PMLR}.
\newblock
\begin{APACrefURL} \url{https://proceedings.mlr.press/v202/kim23g.html} \end{APACrefURL}
\PrintBackRefs{\CurrentBib}

\bibitem [\protect \citeauthoryear {%
Kingma%
\ \BBA {} Ba%
}{%
Kingma%
\ \BBA {} Ba%
}{%
{\protect \APACyear {2015}}%
}]{%
kingma2014adam}
\APACinsertmetastar {%
kingma2014adam}%
\begin{APACrefauthors}%
Kingma, D\BPBI P.%
\BCBT {}\ \BBA {} Ba, J.%
\end{APACrefauthors}%
\unskip\
\newblock
\APACrefYearMonthDay{2015}{}{}.
\newblock
{\BBOQ}\APACrefatitle {Adam: {A} Method for Stochastic Optimization} {Adam: {A} method for stochastic optimization}.{\BBCQ}
\newblock
\BIn{} Y.~Bengio\ \BBA {} Y.~LeCun\ (\BEDS), \APACrefbtitle {3rd International Conference on Learning Representations, {ICLR} 2015, San Diego, CA, USA, May 7-9, 2015, Conference Track Proceedings.} {3rd international conference on learning representations, {ICLR} 2015, san diego, ca, usa, may 7-9, 2015, conference track proceedings.}
\newblock
\begin{APACrefURL} \url{http://arxiv.org/abs/1412.6980} \end{APACrefURL}
\PrintBackRefs{\CurrentBib}

\bibitem [\protect \citeauthoryear {%
Kingma%
\ \BBA {} Welling%
}{%
Kingma%
\ \BBA {} Welling%
}{%
{\protect \APACyear {2014}}%
}]{%
kingma2013auto}
\APACinsertmetastar {%
kingma2013auto}%
\begin{APACrefauthors}%
Kingma, D\BPBI P.%
\BCBT {}\ \BBA {} Welling, M.%
\end{APACrefauthors}%
\unskip\
\newblock
\APACrefYearMonthDay{2014}{}{}.
\newblock
{\BBOQ}\APACrefatitle {Auto-Encoding Variational Bayes} {Auto-encoding variational bayes}.{\BBCQ}
\newblock
\BIn{} Y.~Bengio\ \BBA {} Y.~LeCun\ (\BEDS), \APACrefbtitle {2nd International Conference on Learning Representations, {ICLR} 2014, Banff, AB, Canada, April 14-16, 2014, Conference Track Proceedings.} {2nd international conference on learning representations, {ICLR} 2014, banff, ab, canada, april 14-16, 2014, conference track proceedings.}
\newblock
\begin{APACrefURL} \url{http://arxiv.org/abs/1312.6114} \end{APACrefURL}
\PrintBackRefs{\CurrentBib}

\bibitem [\protect \citeauthoryear {%
Koh%
\ \protect \BOthers {.}}{%
Koh%
\ \protect \BOthers {.}}{%
{\protect \APACyear {2020}}%
}]{%
Koh2020}
\APACinsertmetastar {%
Koh2020}%
\begin{APACrefauthors}%
Koh, P\BPBI W.%
, Nguyen, T.%
, Tang, Y\BPBI S.%
, Mussmann, S.%
, Pierson, E.%
, Kim, B.%
\BCBL {}\ \BBA {} Liang, P.%
\end{APACrefauthors}%
\unskip\
\newblock
\APACrefYearMonthDay{2020}{}{}.
\newblock
{\BBOQ}\APACrefatitle {Concept Bottleneck Models} {Concept bottleneck models}.{\BBCQ}
\newblock
\BIn{} H\BPBI D.~III\ \BBA {} A.~Singh\ (\BEDS), \APACrefbtitle {Proceedings of the 37th International Conference on Machine Learning} {Proceedings of the 37th international conference on machine learning}\ (\BVOL~119, \BPGS\ 5338--5348).
\newblock
\APACaddressPublisher{Virtual}{PMLR}.
\newblock
\begin{APACrefURL} \url{https://proceedings.mlr.press/v119/koh20a.html} \end{APACrefURL}
\PrintBackRefs{\CurrentBib}

\bibitem [\protect \citeauthoryear {%
Kraft%
}{%
Kraft%
}{%
{\protect \APACyear {1988}}%
}]{%
kraft1988software}
\APACinsertmetastar {%
kraft1988software}%
\begin{APACrefauthors}%
Kraft, D.%
\end{APACrefauthors}%
\unskip\
\newblock
\APACrefYearMonthDay{1988}{}{}.
\newblock
{\BBOQ}\APACrefatitle {A software package for sequential quadratic programming} {A software package for sequential quadratic programming}.{\BBCQ}
\newblock
\APACjournalVolNumPages{Forschungsbericht- Deutsche Forschungs- und Versuchsanstalt fur Luft- und Raumfahrt}{}{}{}.
\PrintBackRefs{\CurrentBib}

\bibitem [\protect \citeauthoryear {%
Krizhevsky%
, Hinton%
\BCBL {}\ \protect \BOthers {.}}{%
Krizhevsky%
\ \protect \BOthers {.}}{%
{\protect \APACyear {2009}}%
}]{%
krizhevsky2009learning}
\APACinsertmetastar {%
krizhevsky2009learning}%
\begin{APACrefauthors}%
Krizhevsky, A.%
, Hinton, G.%
\BCBL {}\ \BOthersPeriod {.}\end{APACrefauthors}%
\unskip\
\newblock
\APACrefYearMonthDay{2009}{}{}.
\newblock
{\BBOQ}\APACrefatitle {Learning multiple layers of features from tiny images} {Learning multiple layers of features from tiny images}.{\BBCQ}
\newblock

\PrintBackRefs{\CurrentBib}

\bibitem [\protect \citeauthoryear {%
A.~Kumar%
, Liang%
\BCBL {}\ \BBA {} Ma%
}{%
A.~Kumar%
\ \protect \BOthers {.}}{%
{\protect \APACyear {2019}}%
}]{%
kumar2019verified}
\APACinsertmetastar {%
kumar2019verified}%
\begin{APACrefauthors}%
Kumar, A.%
, Liang, P\BPBI S.%
\BCBL {}\ \BBA {} Ma, T.%
\end{APACrefauthors}%
\unskip\
\newblock
\APACrefYearMonthDay{2019}{}{}.
\newblock
{\BBOQ}\APACrefatitle {Verified uncertainty calibration} {Verified uncertainty calibration}.{\BBCQ}
\newblock
\APACjournalVolNumPages{Advances in Neural Information Processing Systems}{32}{}{}.
\PrintBackRefs{\CurrentBib}

\bibitem [\protect \citeauthoryear {%
N.~Kumar%
, Berg%
, Belhumeur%
\BCBL {}\ \BBA {} Nayar%
}{%
N.~Kumar%
\ \protect \BOthers {.}}{%
{\protect \APACyear {2009}}%
}]{%
Kumar2009}
\APACinsertmetastar {%
Kumar2009}%
\begin{APACrefauthors}%
Kumar, N.%
, Berg, A\BPBI C.%
, Belhumeur, P\BPBI N.%
\BCBL {}\ \BBA {} Nayar, S\BPBI K.%
\end{APACrefauthors}%
\unskip\
\newblock
\APACrefYearMonthDay{2009}{}{}.
\newblock
{\BBOQ}\APACrefatitle {Attribute and simile classifiers for face verification} {Attribute and simile classifiers for face verification}.{\BBCQ}
\newblock
\BIn{} \APACrefbtitle {2009 IEEE 12th International Conference on Computer Vision} {2009 ieee 12th international conference on computer vision}\ (\BPGS\ 365--372).
\newblock
\APACaddressPublisher{Kyoto, Japan}{IEEE}.
\newblock
\begin{APACrefURL} \url{https://doi.org/10.1109/ICCV.2009.5459250} \end{APACrefURL}
\PrintBackRefs{\CurrentBib}

\bibitem [\protect \citeauthoryear {%
Lampert%
, Nickisch%
\BCBL {}\ \BBA {} Harmeling%
}{%
Lampert%
\ \protect \BOthers {.}}{%
{\protect \APACyear {2009}}%
}]{%
Lampert2009}
\APACinsertmetastar {%
Lampert2009}%
\begin{APACrefauthors}%
Lampert, C\BPBI H.%
, Nickisch, H.%
\BCBL {}\ \BBA {} Harmeling, S.%
\end{APACrefauthors}%
\unskip\
\newblock
\APACrefYearMonthDay{2009}{}{}.
\newblock
{\BBOQ}\APACrefatitle {Learning to detect unseen object classes by between-class attribute transfer} {Learning to detect unseen object classes by between-class attribute transfer}.{\BBCQ}
\newblock
\BIn{} \APACrefbtitle {2009 {IEEE} Conference on Computer Vision and Pattern Recognition.} {2009 {IEEE} conference on computer vision and pattern recognition.}
\newblock
\APACaddressPublisher{Miami, FL, USA}{{IEEE}}.
\newblock
\begin{APACrefURL} \url{https://doi.org/10.1109/CVPR.2009.5206594} \end{APACrefURL}
\PrintBackRefs{\CurrentBib}

\bibitem [\protect \citeauthoryear {%
Leino%
, Sen%
, Datta%
, Fredrikson%
\BCBL {}\ \BBA {} Li%
}{%
Leino%
\ \protect \BOthers {.}}{%
{\protect \APACyear {2018}}%
}]{%
Leino2018}
\APACinsertmetastar {%
Leino2018}%
\begin{APACrefauthors}%
Leino, K.%
, Sen, S.%
, Datta, A.%
, Fredrikson, M.%
\BCBL {}\ \BBA {} Li, L.%
\end{APACrefauthors}%
\unskip\
\newblock
\APACrefYearMonthDay{2018}{}{}.
\newblock
{\BBOQ}\APACrefatitle {Influence-Directed Explanations for Deep Convolutional Networks} {Influence-directed explanations for deep convolutional networks}.{\BBCQ}
\newblock
\BIn{} \APACrefbtitle {2018 {IEEE} International Test Conference ({ITC}).} {2018 {IEEE} international test conference ({ITC}).}
\newblock
\APACaddressPublisher{}{{IEEE}}.
\newblock
\begin{APACrefURL} \url{https://doi.org/10.1109/test.2018.8624792} \end{APACrefURL}
\PrintBackRefs{\CurrentBib}

\bibitem [\protect \citeauthoryear {%
Lipton%
}{%
Lipton%
}{%
{\protect \APACyear {2016}}%
}]{%
liptonMythosModelInterpretability2016}
\APACinsertmetastar {%
liptonMythosModelInterpretability2016}%
\begin{APACrefauthors}%
Lipton, Z\BPBI C.%
\end{APACrefauthors}%
\unskip\
\newblock
\APACrefYearMonthDay{2016}{{\APACmonth{06}}}{}.
\newblock
{\BBOQ}\APACrefatitle {The {{Mythos}} of {{Model Interpretability}}} {The {{Mythos}} of {{Model Interpretability}}}.{\BBCQ}
\newblock
\APACjournalVolNumPages{Communications of the ACM}{61}{10}{35--43}.
\newblock
\begin{APACrefDOI} \doi{10.48550/arxiv.1606.03490} \end{APACrefDOI}
\PrintBackRefs{\CurrentBib}

\bibitem [\protect \citeauthoryear {%
Maddison%
, Mnih%
\BCBL {}\ \BBA {} Teh%
}{%
Maddison%
\ \protect \BOthers {.}}{%
{\protect \APACyear {2017}}%
}]{%
concrete}
\APACinsertmetastar {%
concrete}%
\begin{APACrefauthors}%
Maddison, C\BPBI J.%
, Mnih, A.%
\BCBL {}\ \BBA {} Teh, Y\BPBI W.%
\end{APACrefauthors}%
\unskip\
\newblock
\APACrefYearMonthDay{2017}{}{}.
\newblock
{\BBOQ}\APACrefatitle {The Concrete Distribution: {A} Continuous Relaxation of Discrete Random Variables} {The concrete distribution: {A} continuous relaxation of discrete random variables}.{\BBCQ}
\newblock
\BIn{} \APACrefbtitle {5th International Conference on Learning Representations, {ICLR} 2017, Toulon, France, April 24-26, 2017, Conference Track Proceedings.} {5th international conference on learning representations, {ICLR} 2017, toulon, france, april 24-26, 2017, conference track proceedings.}
\newblock
\APACaddressPublisher{}{OpenReview.net}.
\newblock
\begin{APACrefURL} \url{https://openreview.net/forum?id=S1jE5L5gl} \end{APACrefURL}
\PrintBackRefs{\CurrentBib}

\bibitem [\protect \citeauthoryear {%
Mahinpei%
, Clark%
, Lage%
, Doshi-Velez%
\BCBL {}\ \BBA {} Pan%
}{%
Mahinpei%
\ \protect \BOthers {.}}{%
{\protect \APACyear {2021}}%
}]{%
Mahinpei2021}
\APACinsertmetastar {%
Mahinpei2021}%
\begin{APACrefauthors}%
Mahinpei, A.%
, Clark, J.%
, Lage, I.%
, Doshi-Velez, F.%
\BCBL {}\ \BBA {} Pan, W.%
\end{APACrefauthors}%
\unskip\
\newblock
\APACrefYearMonthDay{2021}{}{}.
\newblock
\APACrefbtitle {Promises and Pitfalls of Black-Box Concept Learning Models.} {Promises and pitfalls of black-box concept learning models.}
\newblock
\begin{APACrefURL} \url{https://doi.org/10.48550/arXiv.2106.13314} \end{APACrefURL}
\newblock
\APACrefnote{\textit{arXiv:2106.13314}}
\PrintBackRefs{\CurrentBib}

\bibitem [\protect \citeauthoryear {%
Marcinkevi{\v{c}}s%
, Laguna%
, Vandenhirtz%
\BCBL {}\ \BBA {} Vogt%
}{%
Marcinkevi{\v{c}}s%
\ \protect \BOthers {.}}{%
{\protect \APACyear {2024}}%
}]{%
marcinkevivcs2024beyond}
\APACinsertmetastar {%
marcinkevivcs2024beyond}%
\begin{APACrefauthors}%
Marcinkevi{\v{c}}s, R.%
, Laguna, S.%
, Vandenhirtz, M.%
\BCBL {}\ \BBA {} Vogt, J\BPBI E.%
\end{APACrefauthors}%
\unskip\
\newblock
\APACrefYearMonthDay{2024}{}{}.
\newblock
{\BBOQ}\APACrefatitle {Beyond Concept Bottleneck Models: How to Make Black Boxes Intervenable?} {Beyond concept bottleneck models: How to make black boxes intervenable?}{\BBCQ}
\newblock
\BIn{} \APACrefbtitle {Advances in neural information processing systems} {Advances in neural information processing systems}\ (\BVOL~37).
\PrintBackRefs{\CurrentBib}

\bibitem [\protect \citeauthoryear {%
Marcinkevičs%
\ \protect \BOthers {.}}{%
Marcinkevičs%
\ \protect \BOthers {.}}{%
{\protect \APACyear {2024}}%
}]{%
Marcinkevics2023}
\APACinsertmetastar {%
Marcinkevics2023}%
\begin{APACrefauthors}%
Marcinkevičs, R.%
, {Reis Wolfertstetter}, P.%
, Klimiene, U.%
, Chin-Cheong, K.%
, Paschke, A.%
, Zerres, J.%
\BDBL {}Vogt, J\BPBI E.%
\end{APACrefauthors}%
\unskip\
\newblock
\APACrefYearMonthDay{2024}{}{}.
\newblock
{\BBOQ}\APACrefatitle {Interpretable and intervenable ultrasonography-based machine learning models for pediatric appendicitis} {Interpretable and intervenable ultrasonography-based machine learning models for pediatric appendicitis}.{\BBCQ}
\newblock
\APACjournalVolNumPages{Medical Image Analysis}{91}{}{103042}.
\newblock
\begin{APACrefURL} \url{https://www.sciencedirect.com/science/article/pii/S136184152300302X} \end{APACrefURL}
\PrintBackRefs{\CurrentBib}

\bibitem [\protect \citeauthoryear {%
Margeloiu%
\ \protect \BOthers {.}}{%
Margeloiu%
\ \protect \BOthers {.}}{%
{\protect \APACyear {2021}}%
}]{%
Margeloiu2021}
\APACinsertmetastar {%
Margeloiu2021}%
\begin{APACrefauthors}%
Margeloiu, A.%
, Ashman, M.%
, Bhatt, U.%
, Chen, Y.%
, Jamnik, M.%
\BCBL {}\ \BBA {} Weller, A.%
\end{APACrefauthors}%
\unskip\
\newblock
\APACrefYearMonthDay{2021}{}{}.
\newblock
\APACrefbtitle {Do Concept Bottleneck Models Learn as Intended?} {Do concept bottleneck models learn as intended?}
\newblock
\begin{APACrefURL} \url{https://doi.org/10.48550/arXiv.2105.04289} \end{APACrefURL}
\newblock
\APACrefnote{\textit{arXiv:2105.04289}}
\PrintBackRefs{\CurrentBib}

\bibitem [\protect \citeauthoryear {%
Monteiro%
\ \protect \BOthers {.}}{%
Monteiro%
\ \protect \BOthers {.}}{%
{\protect \APACyear {2020}}%
}]{%
monteiro2020stochastic}
\APACinsertmetastar {%
monteiro2020stochastic}%
\begin{APACrefauthors}%
Monteiro, M.%
, Le~Folgoc, L.%
, Coelho~de Castro, D.%
, Pawlowski, N.%
, Marques, B.%
, Kamnitsas, K.%
\BDBL {}Glocker, B.%
\end{APACrefauthors}%
\unskip\
\newblock
\APACrefYearMonthDay{2020}{}{}.
\newblock
{\BBOQ}\APACrefatitle {Stochastic segmentation networks: Modelling spatially correlated aleatoric uncertainty} {Stochastic segmentation networks: Modelling spatially correlated aleatoric uncertainty}.{\BBCQ}
\newblock
\BIn{} \APACrefbtitle {Advances in neural information processing systems} {Advances in neural information processing systems}\ (\BVOL~33, \BPGS\ 12756--12767).
\PrintBackRefs{\CurrentBib}

\bibitem [\protect \citeauthoryear {%
Naeini%
, Cooper%
\BCBL {}\ \BBA {} Hauskrecht%
}{%
Naeini%
\ \protect \BOthers {.}}{%
{\protect \APACyear {2015}}%
}]{%
naeini2015obtaining}
\APACinsertmetastar {%
naeini2015obtaining}%
\begin{APACrefauthors}%
Naeini, M\BPBI P.%
, Cooper, G.%
\BCBL {}\ \BBA {} Hauskrecht, M.%
\end{APACrefauthors}%
\unskip\
\newblock
\APACrefYearMonthDay{2015}{}{}.
\newblock
{\BBOQ}\APACrefatitle {Obtaining well calibrated probabilities using bayesian binning} {Obtaining well calibrated probabilities using bayesian binning}.{\BBCQ}
\newblock
\BIn{} \APACrefbtitle {Proceedings of the AAAI conference on artificial intelligence} {Proceedings of the aaai conference on artificial intelligence}\ (\BVOL~29).
\PrintBackRefs{\CurrentBib}

\bibitem [\protect \citeauthoryear {%
Neal%
}{%
Neal%
}{%
{\protect \APACyear {1995}}%
}]{%
neal2012bayesian}
\APACinsertmetastar {%
neal2012bayesian}%
\begin{APACrefauthors}%
Neal, R\BPBI M.%
\end{APACrefauthors}%
\unskip\
\newblock
\APACrefYear{1995}.
\unskip\
\newblock
\APACrefbtitle {Bayesian learning for neural networks} {Bayesian learning for neural networks}\ \APACtypeAddressSchool {\BPhD}{}{University of Toronto, Canada}.
\unskip\
\newblock
\begin{APACrefURL} \url{https://librarysearch.library.utoronto.ca/permalink/01UTORONTO\_INST/14bjeso/alma991106438365706196} \end{APACrefURL}
\PrintBackRefs{\CurrentBib}

\bibitem [\protect \citeauthoryear {%
Oikarinen%
, Das%
, Nguyen%
\BCBL {}\ \BBA {} Weng%
}{%
Oikarinen%
\ \protect \BOthers {.}}{%
{\protect \APACyear {2023}}%
}]{%
oikarinen2023label}
\APACinsertmetastar {%
oikarinen2023label}%
\begin{APACrefauthors}%
Oikarinen, T.%
, Das, S.%
, Nguyen, L\BPBI M.%
\BCBL {}\ \BBA {} Weng, T\BHBI W.%
\end{APACrefauthors}%
\unskip\
\newblock
\APACrefYearMonthDay{2023}{}{}.
\newblock
{\BBOQ}\APACrefatitle {Label-free Concept Bottleneck Models} {Label-free concept bottleneck models}.{\BBCQ}
\newblock
\BIn{} \APACrefbtitle {The 11th International Conference on Learning Representations.} {The 11th international conference on learning representations.}
\newblock
\begin{APACrefURL} \url{https://openreview.net/forum?id=FlCg47MNvBA} \end{APACrefURL}
\PrintBackRefs{\CurrentBib}

\bibitem [\protect \citeauthoryear {%
Panousis%
, Ienco%
\BCBL {}\ \BBA {} Marcos%
}{%
Panousis%
\ \protect \BOthers {.}}{%
{\protect \APACyear {2023}}%
}]{%
panousis2023sparse}
\APACinsertmetastar {%
panousis2023sparse}%
\begin{APACrefauthors}%
Panousis, K\BPBI P.%
, Ienco, D.%
\BCBL {}\ \BBA {} Marcos, D.%
\end{APACrefauthors}%
\unskip\
\newblock
\APACrefYearMonthDay{2023}{}{}.
\newblock
{\BBOQ}\APACrefatitle {Sparse linear concept discovery models} {Sparse linear concept discovery models}.{\BBCQ}
\newblock
\BIn{} \APACrefbtitle {Proceedings of the IEEE/CVF International Conference on Computer Vision} {Proceedings of the ieee/cvf international conference on computer vision}\ (\BPGS\ 2767--2771).
\PrintBackRefs{\CurrentBib}

\bibitem [\protect \citeauthoryear {%
Panousis%
, Ienco%
\BCBL {}\ \BBA {} Marcos%
}{%
Panousis%
\ \protect \BOthers {.}}{%
{\protect \APACyear {2024}}%
}]{%
panousis2024coarse}
\APACinsertmetastar {%
panousis2024coarse}%
\begin{APACrefauthors}%
Panousis, K\BPBI P.%
, Ienco, D.%
\BCBL {}\ \BBA {} Marcos, D.%
\end{APACrefauthors}%
\unskip\
\newblock
\APACrefYearMonthDay{2024}{}{}.
\newblock
{\BBOQ}\APACrefatitle {Coarse-to-Fine Concept Bottleneck Models} {Coarse-to-fine concept bottleneck models}.{\BBCQ}
\newblock
\BIn{} \APACrefbtitle {NeurIPS 2024-38th Annual Conference on Neural Information Processing Systems.} {Neurips 2024-38th annual conference on neural information processing systems.}
\PrintBackRefs{\CurrentBib}

\bibitem [\protect \citeauthoryear {%
Radford%
\ \protect \BOthers {.}}{%
Radford%
\ \protect \BOthers {.}}{%
{\protect \APACyear {2021}}%
}]{%
radford2021learning}
\APACinsertmetastar {%
radford2021learning}%
\begin{APACrefauthors}%
Radford, A.%
, Kim, J\BPBI W.%
, Hallacy, C.%
, Ramesh, A.%
, Goh, G.%
, Agarwal, S.%
\BDBL {}others%
\end{APACrefauthors}%
\unskip\
\newblock
\APACrefYearMonthDay{2021}{}{}.
\newblock
{\BBOQ}\APACrefatitle {Learning transferable visual models from natural language supervision} {Learning transferable visual models from natural language supervision}.{\BBCQ}
\newblock
\BIn{} \APACrefbtitle {International conference on machine learning} {International conference on machine learning}\ (\BPGS\ 8748--8763).
\PrintBackRefs{\CurrentBib}

\bibitem [\protect \citeauthoryear {%
Sheth%
, Rahman%
, Sevyeri%
, Havaei%
\BCBL {}\ \BBA {} Kahou%
}{%
Sheth%
\ \protect \BOthers {.}}{%
{\protect \APACyear {2022}}%
}]{%
Sheth2022}
\APACinsertmetastar {%
Sheth2022}%
\begin{APACrefauthors}%
Sheth, I.%
, Rahman, A\BPBI A.%
, Sevyeri, L\BPBI R.%
, Havaei, M.%
\BCBL {}\ \BBA {} Kahou, S\BPBI E.%
\end{APACrefauthors}%
\unskip\
\newblock
\APACrefYearMonthDay{2022}{}{}.
\newblock
{\BBOQ}\APACrefatitle {Learning from uncertain concepts via test time interventions} {Learning from uncertain concepts via test time interventions}.{\BBCQ}
\newblock
\BIn{} \APACrefbtitle {Workshop on Trustworthy and Socially Responsible Machine Learning, NeurIPS 2022.} {Workshop on trustworthy and socially responsible machine learning, neurips 2022.}
\newblock
\begin{APACrefURL} \url{https://openreview.net/forum?id=WVe3vok8Cc3} \end{APACrefURL}
\PrintBackRefs{\CurrentBib}

\bibitem [\protect \citeauthoryear {%
Shin%
, Jo%
, Ahn%
\BCBL {}\ \BBA {} Lee%
}{%
Shin%
\ \protect \BOthers {.}}{%
{\protect \APACyear {2023}}%
}]{%
Shin2023}
\APACinsertmetastar {%
Shin2023}%
\begin{APACrefauthors}%
Shin, S.%
, Jo, Y.%
, Ahn, S.%
\BCBL {}\ \BBA {} Lee, N.%
\end{APACrefauthors}%
\unskip\
\newblock
\APACrefYearMonthDay{2023}{}{}.
\newblock
{\BBOQ}\APACrefatitle {A Closer Look at the Intervention Procedure of Concept Bottleneck Models} {A closer look at the intervention procedure of concept bottleneck models}.{\BBCQ}
\newblock
\BIn{} A.~Krause, E.~Brunskill, K.~Cho, B.~Engelhardt, S.~Sabato\BCBL {}\ \BBA {} J.~Scarlett\ (\BEDS), \APACrefbtitle {Proceedings of the 40th International Conference on Machine Learning} {Proceedings of the 40th international conference on machine learning}\ (\BVOL~202, \BPGS\ 31504--31520).
\newblock
\APACaddressPublisher{}{PMLR}.
\newblock
\begin{APACrefURL} \url{https://proceedings.mlr.press/v202/shin23a.html} \end{APACrefURL}
\PrintBackRefs{\CurrentBib}

\bibitem [\protect \citeauthoryear {%
Silvey%
}{%
Silvey%
}{%
{\protect \APACyear {1975}}%
}]{%
silvey1975statistical}
\APACinsertmetastar {%
silvey1975statistical}%
\begin{APACrefauthors}%
Silvey, S.%
\end{APACrefauthors}%
\unskip\
\newblock
\APACrefYear{1975}.
\newblock
\APACrefbtitle {Statistical Inference} {Statistical inference}.
\newblock
\APACaddressPublisher{}{Taylor \& Francis}.
\newblock
\begin{APACrefURL} \url{https://books.google.ch/books?id=qIKLejbVMf4C} \end{APACrefURL}
\PrintBackRefs{\CurrentBib}

\bibitem [\protect \citeauthoryear {%
Singhi%
, Kim%
, Roth%
\BCBL {}\ \BBA {} Akata%
}{%
Singhi%
\ \protect \BOthers {.}}{%
{\protect \APACyear {2024}}%
}]{%
singhi2024improving}
\APACinsertmetastar {%
singhi2024improving}%
\begin{APACrefauthors}%
Singhi, N.%
, Kim, J\BPBI M.%
, Roth, K.%
\BCBL {}\ \BBA {} Akata, Z.%
\end{APACrefauthors}%
\unskip\
\newblock
\APACrefYearMonthDay{2024}{}{}.
\newblock
{\BBOQ}\APACrefatitle {Improving Intervention Efficacy via Concept Realignment in Concept Bottleneck Models} {Improving intervention efficacy via concept realignment in concept bottleneck models}.{\BBCQ}
\newblock
\APACjournalVolNumPages{arXiv preprint arXiv:2405.01531}{}{}{}.
\PrintBackRefs{\CurrentBib}

\bibitem [\protect \citeauthoryear {%
Steinmann%
, Stammer%
, Friedrich%
\BCBL {}\ \BBA {} Kersting%
}{%
Steinmann%
\ \protect \BOthers {.}}{%
{\protect \APACyear {2023}}%
}]{%
Steinmann2023}
\APACinsertmetastar {%
Steinmann2023}%
\begin{APACrefauthors}%
Steinmann, D.%
, Stammer, W.%
, Friedrich, F.%
\BCBL {}\ \BBA {} Kersting, K.%
\end{APACrefauthors}%
\unskip\
\newblock
\APACrefYearMonthDay{2023}{}{}.
\newblock
\APACrefbtitle {Learning to Intervene on Concept Bottlenecks.} {Learning to intervene on concept bottlenecks.}
\newblock
\begin{APACrefURL} \url{https://doi.org/10.48550/arXiv.2308.13453} \end{APACrefURL}
\newblock
\APACrefnote{\textit{arXiv:2308.13453}}
\PrintBackRefs{\CurrentBib}

\bibitem [\protect \citeauthoryear {%
Wah%
, Branson%
, Welinder%
, Perona%
\BCBL {}\ \BBA {} Belongie%
}{%
Wah%
\ \protect \BOthers {.}}{%
{\protect \APACyear {2011}}%
}]{%
wah2011caltech}
\APACinsertmetastar {%
wah2011caltech}%
\begin{APACrefauthors}%
Wah, C.%
, Branson, S.%
, Welinder, P.%
, Perona, P.%
\BCBL {}\ \BBA {} Belongie, S.%
\end{APACrefauthors}%
\unskip\
\newblock
\APACrefYearMonthDay{2011}{}{}.
\newblock
{\BBOQ}\APACrefatitle {The caltech-ucsd birds-200-2011 dataset} {The caltech-ucsd birds-200-2011 dataset}.{\BBCQ}
\newblock

\PrintBackRefs{\CurrentBib}

\bibitem [\protect \citeauthoryear {%
Yuksekgonul%
, Wang%
\BCBL {}\ \BBA {} Zou%
}{%
Yuksekgonul%
\ \protect \BOthers {.}}{%
{\protect \APACyear {2023}}%
}]{%
Yuksekgonul2023}
\APACinsertmetastar {%
Yuksekgonul2023}%
\begin{APACrefauthors}%
Yuksekgonul, M.%
, Wang, M.%
\BCBL {}\ \BBA {} Zou, J.%
\end{APACrefauthors}%
\unskip\
\newblock
\APACrefYearMonthDay{2023}{}{}.
\newblock
{\BBOQ}\APACrefatitle {Post-hoc Concept Bottleneck Models} {Post-hoc concept bottleneck models}.{\BBCQ}
\newblock
\BIn{} \APACrefbtitle {The 11th International Conference on Learning Representations.} {The 11th international conference on learning representations.}
\newblock
\begin{APACrefURL} \url{https://openreview.net/forum?id=nA5AZ8CEyow} \end{APACrefURL}
\PrintBackRefs{\CurrentBib}

\end{thebibliography}
